%% file: main.tex
\documentclass[lettersize,journal]{IEEEtran}
\usepackage{amsmath,amsfonts}
\usepackage{algorithmic}
\usepackage{algorithm}
\usepackage{array}
\usepackage[caption=false,font=normalsize,labelfont=sf,textfont=sf]{subfig}
\usepackage{textcomp}
\usepackage{stfloats}
\usepackage{url}
\usepackage{verbatim}
\usepackage{graphicx}
\usepackage{cite}
\usepackage{booktabs}

\usepackage{xcolor}
\usepackage{color}

\usepackage{multirow}

\usepackage[numbers,sort&compress]{natbib}

\usepackage[figuresleft]{rotating}

\begin{document}

\title{Milestones in Autonomous Driving and Intelligent Vehicles Part \uppercase\expandafter{\romannumeral1}: Control, Computing System Design, Communication, HD Map, Testing, and Human Behaviors}

\author{Long Chen, ~\IEEEmembership{Senior Member,~IEEE,} Yuchen Li, Chao Huang, Yang Xing, 

Daxin Tian, ~\IEEEmembership{Senior Member,~IEEE,} Li Li,~\IEEEmembership{Fellow,~IEEE,} Zhongxu Hu, Siyu Teng, 

Chen Lv, ~\IEEEmembership{Senior Member,~IEEE,} Jinjun Wang, Dongpu Cao, ~\IEEEmembership{Senior Member,~IEEE,}

Nanning Zheng,~\IEEEmembership{Fellow,~IEEE} and Fei-Yue Wang,~\IEEEmembership{Fellow,~IEEE} 
\thanks{Manuscript received April 19, 2021; revised August 16, 2021.
(Corresponding author: Fei-Yue Wang.)

This work is supported by the National Natural Science Foundation of China (62006256) and the Key Research and Development Program of Guangzhou (202007050002  202007050004).

Long Chen and Fei-Yue Wang are with the State Key Laboratory of Management and Control for Complex Systems, Institute of Automation, Chinese Academy of Sciences, Beijing, 100190, China, and Long Chen is also with Waytous Ltd. (e-mail: long.chen@ia.ac.cn; feiyue.wang@ia.ac.cn). 

Yuchen Li and Siyu Teng are with BNU-HKBU United International College, Zhuhai, 519087, China and Hong Kong Baptist University, Kowloon, Hong Kong, 999077, China (e-mail: liyuchen2016@hotmail.com; siyuteng@ieee.org).

Chao Huang is with the Department of Industrial and System Engineering, Hong Kong Polytechnical University (e-mail: hchao.huang@polyu.edu.hk).

Yang Xing is with the School of Aerospace, Transport, and Manufacturing, Cranfield
University (e-mail: yang.x@cranfield.ac.uk).

Daxin Tian is with the School of Transportation Science and Engineering, Beihang University (e-mail: dtian@buaa.edu.cn). 

Li Li is with the Department of Automation, Tsinghua University (e-mail: li-li@tsinghua.edu.cn). 

Zhongxu Hu and Chen Lv are with the Department of Mechanical and Aerospace Engineering of Nanyang Technological University (e-mail: Huzhongxu.hu@ntu.edu.sg; lyuchen@ntu.edu.sg). 

Jinjun Wang and Nanning Zheng are with the College of Artificial Intelligence, Xi'an Jiaotong University (e-mail: jinjun@mail.xjtu.edu.cn; nnzheng@mail.xjtu.edu.cn).

Dongpu Cao is with the School of Mechanical Engineering, Tsinghua University (e-mail: dp\_cao2016@163.com).}}

\markboth{Journal of \LaTeX\ Class Files,~Vol.~14, No.~8, August~2021}%
{Shell \MakeLowercase{\textit{et al.}}: A Sample Article Using IEEEtran.cls for IEEE Journals}


\maketitle

\input{body/abstract}
\input{body/introduction}
\input{body/methodology}
\input{body/conclusion}

\input{body/bib}


\section{Biography Section}

\begin{IEEEbiography}[{\includegraphics[width=1in,height=1.25in,clip,keepaspectratio]{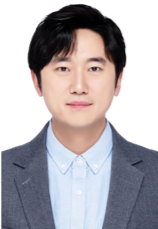}}]{Long Chen} (Senior Member, IEEE) received the Ph.D. degree from Wuhan University in 2013, he is currently a Professor with State Key Laboratory of Management and Control for Complex Systems, Institute of Automation, Chinese Academy of Sciences, Beijing, China. His research interests include autonomous driving, robotics, and artificial intelligence, where he has contributed more than 100 publications. He serves as an Associate Editor for the IEEE Transaction on Intelligent Transportation Systems, the IEEE/CAA Journal of Automatica Sinica, the IEEE Transaction on Intelligent Vehicle and the IEEE Technical Committee on Cyber-Physical Systems.
\end{IEEEbiography}

\begin{IEEEbiography}[{\includegraphics[width=1in,height=1.25in,clip,keepaspectratio]{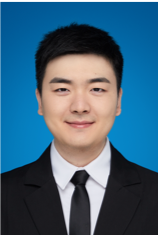}}]{Yuchen Li} received the B.E. degree from the University of Science and Technology Beijing in 2016, and the M.E. degrees from Beihang University in 2020. He is pursuing the Ph.D. degree in Hong Kong Baptist University. He is a intern in Waytous. His research interest covers computer vision, 3D object detection, and autonomous driving.
\end{IEEEbiography}

\begin{IEEEbiography}[{\includegraphics[width=1in,height=1.25in,clip,keepaspectratio]{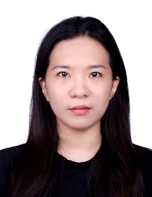}}]{Chao Huang} Chao Huang received her B.S. degree in Control Engineering from the China University of Petroleum, Beijing, and a Ph.D. degree from the University of Wollongong, Australia in 2012 and 2018, respectively. She is currently a research assistant professor at the Department of Industrial and Systems Engineering, The Hong Kong Polytechnic University. Her research interests are human-machine collaboration, fault-tolerant control, mobile robot (EV, UAV), and path planning and control.  She has served on program committees and helped to organize special issues on Sensors, Machines, and Aerospace. She has served as review editor of Frontiers in Mechanical Engineering and associate editor of IEEE Transactions on Intelligent Vehicles.
\end{IEEEbiography}


\begin{IEEEbiography}[{\includegraphics[width=1in,height=1.25in,clip,keepaspectratio]{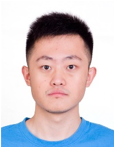}}]{Yang Xing} received his Ph. D. degree from Cranfield University, UK, in 2018. He joined Cranfield University as a Lecturer in Applied AI for Engineering in 2021. Before that, he finished his post-doctoral fellowship at the University of Oxford and Nanyang Technological University, respectively.  His research interests include machine learning, driver behaviours, intelligent multi-agent collaboration, and intelligent/autonomous vehicles. He received the IV2018 Best Workshop/Special Issue Paper Award. Dr Xing serves as an associate editor for IEEE Transactions on Intelligent Vehicle. He was also a Guest Editor for IEEE Internet of Things Journal, and IEEE ITS Magazine, etc. 
\end{IEEEbiography}

\begin{IEEEbiography}[{\includegraphics[width=1in,height=1.25in,clip,keepaspectratio]{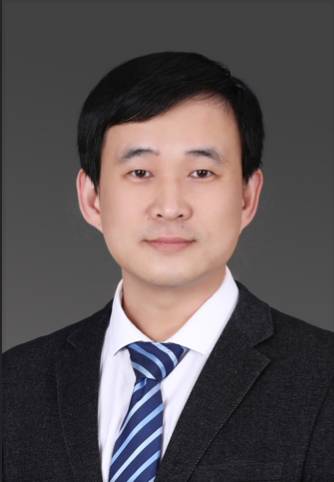}}]{Daxin Tian}
 (Senior Member, IEEE) received the B.S., M.S., and Ph.D. degrees in computer science from Jilin University, Changchun, China, in 2002, 2005, and 2007, respectively. He is currently a professor in the School of Transportation Science and Engineering, Beihang University, Beijing, China. He is an IEEE Intelligent Transportation Systems Society Member and an IEEE Vehicular Technology Society Member. His current research interests include mobile computing, intelligent transportation systems, vehicular ad hoc networks, and swarm intelligent.
\end{IEEEbiography}

\begin{IEEEbiography}[{\includegraphics[width=1in,height=1.25in,clip,keepaspectratio]{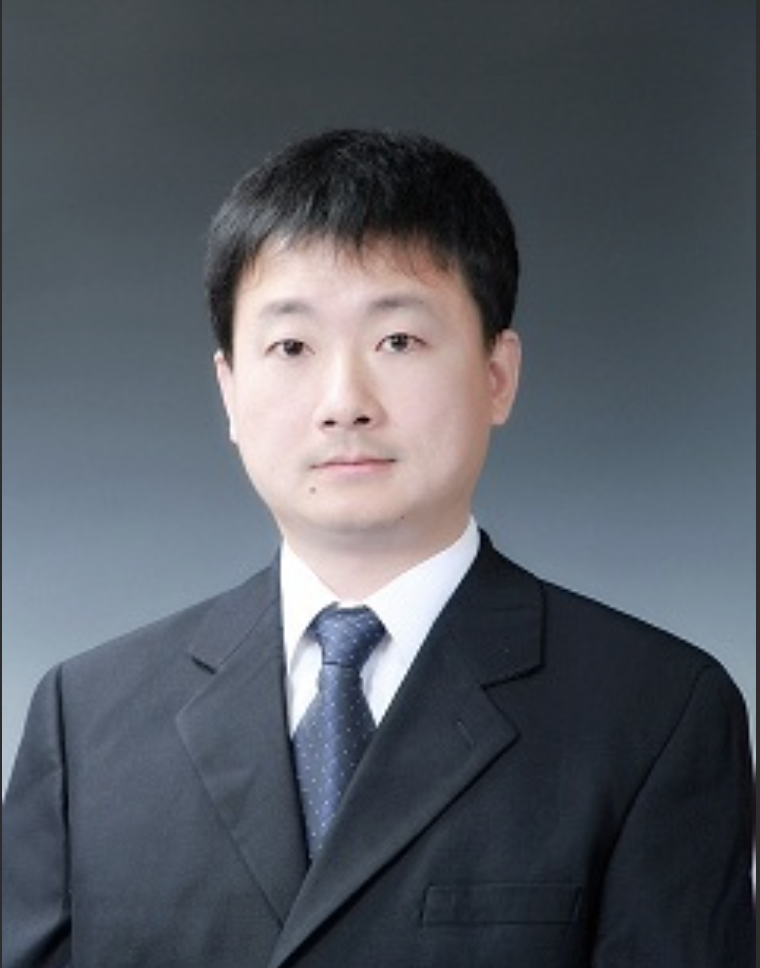}}]{Li Li} (Fellow, IEEE) is a Professor at Tsinghua University. He has been engaged in scientific research in the fields of intelligent transportation and intelligent vehicles. He has published more than 120 SCI search papers as the first or corresponding author. He was a member of the Editorial Advisory Board for Transportation Research Part C: Emerging Technologies, a member of the Editorial Board of Transport Reviews and ACTA Automatica Sinica. He serves as Associate Editors for the IEEE Transactions on Intelligent Transportation Systems and IEEE Transactions on Intelligent Vehicles. He is a Fellow of IEEE and a Fellow of CAA.
\end{IEEEbiography}

\begin{IEEEbiography}[{\includegraphics[width=1in,height=1.25in,clip,keepaspectratio]{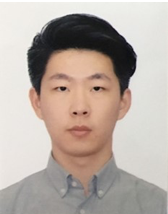}}]{Zhongxu Hu} received a mechatronic Ph.D. degree from the Huazhong University of Science and Technology of China, in 2018. He was a senior engineer at Huawei. He is currently a research fellow within the Department of Mechanical and Aerospace Engineering of Nanyang Technological University in Singapore. His current research interests include human-machine interaction, computer vision, and deep learning applied to driver behavior analysis and autonomous vehicles in multiple scenarios. Dr. Hu served as a Lead Guest Editor for Computational Intelligence and Neuroscience, an Academic Editor/Editorial Board for Automotive Innovation, Journal of Electrical and Electronic Engineering, and is also an active reviewer for IEEE Transactions on Intelligent Transportation Systems, IEEE Transactions on Industrial Electronics, IEEE Intelligent Transportation Systems Magazine, Journal of Intelligent Manufacturing, and Journal of Advanced Transportation et al.
\end{IEEEbiography}



\begin{IEEEbiography}[{\includegraphics[width=1in,height=1.25in,clip,keepaspectratio]{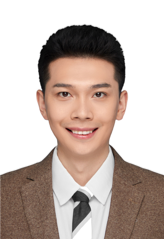}}]{Siyu Teng} received M.S. degree from Jilin University in 2021. Now he is a PhD Student at Department of Computer Science, Faculty of Science, Hong Kong Baptist University. His main interests are parallel planning, end-to-end autonomous driving and interpretable deep learning.
\end{IEEEbiography}

\begin{IEEEbiography}[{\includegraphics[width=1in,height=1.25in,clip,keepaspectratio]{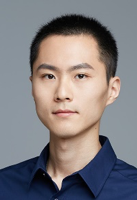}}]{Chen Lv} (Senior Member, IEEE) is a Nanyang Assistant Professor at School of Mechanical and Aerospace Engineering, Nanyang Technological University, Singapore. He received his PhD degree at Department of Automotive Engineering, Tsinghua University, China, with a joint PhD at UC Berkeley. His research focuses on intelligent vehicles, automated driving, and human-machine systems, where he has contributed 2 books, over 100 papers, and obtained 12 granted patents. He serves as Associate Editor for IEEE T-ITS, IEEE TVT, and IEEE T-IV. He received many awards and honors, selectively including the Highly Commended Paper Award of IMechE UK in 2012, Japan NSK Outstanding Mechanical Engineering Paper Award in 2014, Tsinghua University Outstanding Doctoral Thesis Award in 2016, IEEE IV Best Workshop/Special Session Paper Award in 2018, Automotive Innovation Best Paper Award in 2020, the winner of the Waymo Open Dataset Challenges at CVPR 2021, and Machines Young Investigator Award in 2022.
\end{IEEEbiography}

\begin{IEEEbiography}[{\includegraphics[width=1in,height=1.25in,clip,keepaspectratio]{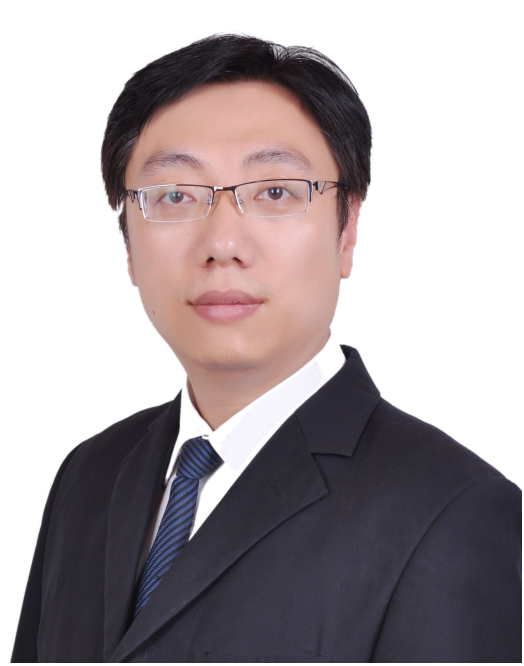}}]{Jinjun Wang} received the B.E. and M.E. degrees from Huazhong University of Science and Technology, China, in 2000 and 2003, respectively, and the Ph.D. degree from Nanyang Technological University, Singapore, in 2006. From 2006 to 2009, he was with NEC Laboratories America, Inc., as a Research Scientist, and from 2010 to 2013, he was with Epson Research and Development, Inc., as a Senior Research Scientist. He is currently a Professor with Xi’an Jiaotong University. His research interests include pattern classification, image/video enhancement and editing, content-based image/video annotation and retrieval, and semantic event detection.
\end{IEEEbiography}

\begin{IEEEbiography}[{\includegraphics[width=1in,height=1.25in,clip,keepaspectratio]{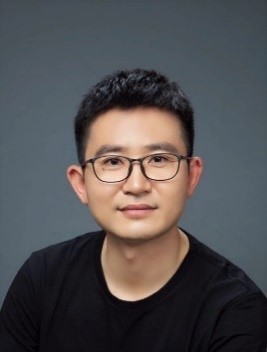}}]{Dongpu Cao} (Senior Member, IEEE) received the Ph.D. degree from Concordia University, Canada, in 2008. He is the Canada Research Chair in Driver Cognition and Automated Driving, and currently an Associate Professor and Director of Waterloo Cognitive Autonomous Driving (CogDrive) Lab at University of Waterloo, Canada. His current research focuses on driver cognition, automated driving and cognitive autonomous driving. He has contributed more than 200 papers and 3 books. He received the SAE Arch T. Colwell Merit Award in 2012, and three Best Paper Awards from the ASME and IEEE conferences. Dr. Cao serves as an Associate Editor for IEEE TRANSACTIONS ON VEHICULAR TECHNOLOGY, IEEE TRANSACTIONS ON INTELLIGENT TRANSPORTATION SYSTEMS, IEEE/ASME TRANSACTIONS ON MECHATRONICS, IEEE TRANSACTIONS ON INDUSTRIAL ELECTRONICS, IEEE/CAA JOURNAL OF AUTOMATICA SINICA and ASME JOURNAL OF DYNAMIC SYSTEMS, MEASUREMENT AND CONTROL. He was a Guest Editor for VEHICLE SYSTEM DYNAMICS and IEEE TRANSACTIONS ON SMC: SYSTEMS. He serves on the SAE Vehicle Dynamics Standards Committee and acts as the Co-Chair of IEEE ITSS Technical Committee on Cooperative Driving.
\end{IEEEbiography}

\begin{IEEEbiography}[{\includegraphics[width=1in,height=1.25in,clip,keepaspectratio]{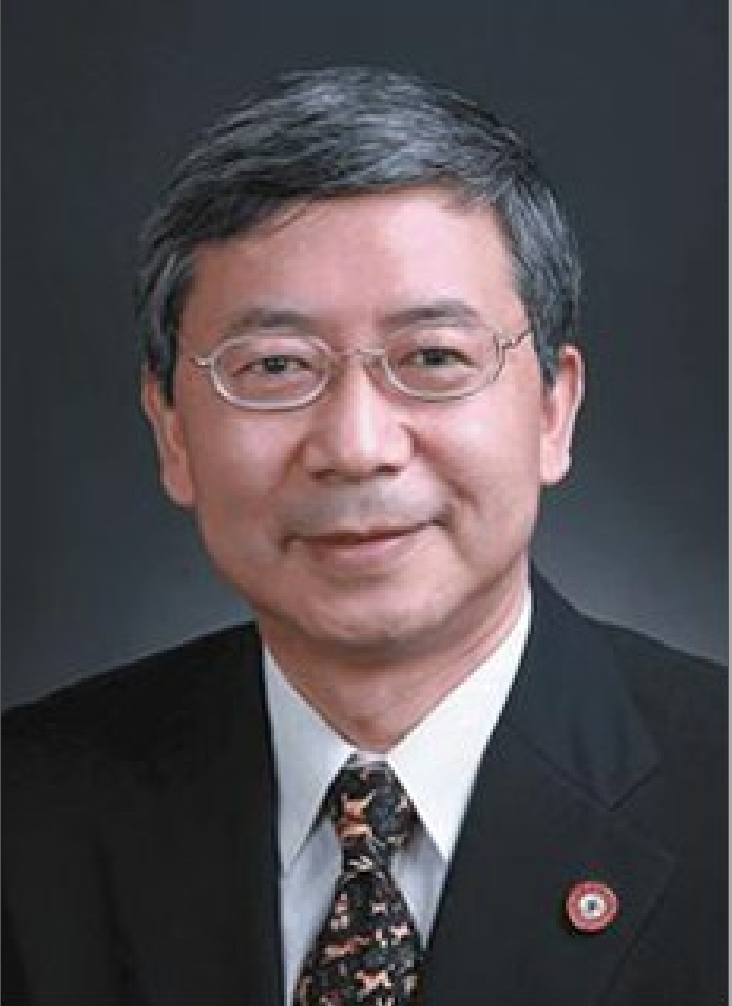}}]{Nanning Zheng} (Fellow, IEEE) graduated from the Department of Electrical Engineering, Xian Jiaotong University, Xian, China, in 1975, and received the M.S. degree in information and control engineering from Xian Jiaotong University in 1981 and the Ph.D. degree in electrical engineering from Keio University, Yokohama, Japan, in 1985. He jointed Xian Jiaotong University in 1975, and he is currently a Professor and the Director of the Institute of Artificial Intelligence and Robotics, Xian Jiaotong University. His research interests include computer vision, pattern recognition and image processing, and hardware implementation of intelligent systems. Dr. Zheng became a member of the Chinese Academy of Engineering in 1999, and he is the Chinese Representative on the Governing Board of the International Association for Pattern Recognition.
\end{IEEEbiography}

\begin{IEEEbiography}[{\includegraphics[width=1in,height=1.25in,clip,keepaspectratio]{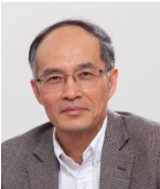}}]{Fei-Yue Wang} (Fellow, IEEE) received the Ph.D. degree in computer and systems engineering from the Rensselaer Polytechnic Institute, Troy, NY, USA, in 1990. He is currently a Professor and the Director of the State Key Laboratory of Intelligent Control and Management of Complex Systems, Institute of Automation, Chinese Academy of Sciences, Beijing, China. He is a member of Sigma Xi and an Elected Fellow of INCOSE, IFAC, ASME, and AAAS. He was the President of the IEEE Intelligent Transportation Systems Society from 2005 to 2007, the Chinese Association for Science and Technology, USA, in 2005, and the American Zhu Kezhen Education Foundation from 2007 to 2008. He is currently the Vice President and the Secretary General of the Chinese Association of Automation. He was the Founding Editor-in-Chief of the International Journal of Intelligent Control and Systems from 1995 to 2000, the Series on Intelligent Control and Intelligent Automation from 1996 to 2004, and the IEEE Transactions on Intelligent Transportation Systems. He was the Editor-in-Chief of the IEEE Intelligent Systems from 2009 to 2011 and the IEEE Transactions on Intelligent Transportation Systems. He is the Editor-in-Chief of the IEEE/CAA Journal of Automatica Sinica.
\end{IEEEbiography}

\vfill

\end{document}

%% file: body/abstract.tex
\begin{abstract}

Interest in autonomous driving (AD) and intelligent vehicles (IVs) is growing at a rapid pace due to the convenience, safety, and economic benefits. Although a number of surveys have reviewed research achievements in this field, they are still limited in specific tasks and lack systematic summaries and research directions in the future. Our work is divided into 3 independent articles and the first part is a Survey of Surveys (SoS) for total technologies of AD and IVs that involves the history, summarizes the milestones, and provides the perspectives, ethics, and future research directions. This is the second part (Part \uppercase\expandafter{\romannumeral1} for this technical survey) to review the development of control, computing system design, communication, High Definition map (HD map), testing, and human behaviors in IVs. In addition, the third part (Part \uppercase\expandafter{\romannumeral2} for this technical survey) is to review the perception and planning sections. The objective of this paper is to involve all the sections of AD, summarize the latest technical milestones, and guide abecedarians to quickly understand the development of AD and IVs. Combining the SoS and Part \uppercase\expandafter{\romannumeral2}, we anticipate that this work will bring novel and diverse insights to researchers and abecedarians, and serve as a bridge between past and future.

\end{abstract}

\begin{IEEEkeywords}
Autonomous Driving, Intelligent Vehicles, Control, Computing System Design, Communication, HD Map, Testing, Human Behaviors, Perception, Planning, Survey of Surveys.
\end{IEEEkeywords}

%% file: body/introduction.tex
\section{Introduction}

\IEEEPARstart{A}{utonomous} driving (AD) and intelligent vehicles (IVs) have recently attracted significant attention from academia as well as industry because of a range of potential benefits. Surveys on AD and IVs occupy an essential position in gathering research achievements, generalizing entire technology development, and forecasting future trends. However, a large majority of surveys only focus on specific tasks and lack systematic summaries and research directions in the future. As a result, they may have a negative impact on conducting research for abecedarians. Our work consists of 3 independent articles including a Survey of Surveys (SoS) \cite{SOS} and two surveys on crucial technologies of AD and IVs. Here is the second part (Part \uppercase\expandafter{\romannumeral1} of the survey) to systematically review the development of control, computing system design, communication, High Definition map (HD map), testing, and human behaviors in IVs. Combining with the SoS and the third part (Part \uppercase\expandafter{\romannumeral2} of the survey on perception and planning), we expect that our work can be considered as a bridge between past and future for AD and IVs.

We provide the common definitions of AD and IV. AD refers to the technology of making the vehicle capable of sensing the surrounding environment and operating without human involvement. IV refers to a vehicle that owns the above technology to enable partial or fully automated functions. Depending on the different tasks in AD, we divide them into 8 sections, perception, planning, control, computing system design, communication, HD map, testing, and human behaviors in IVs. The perception section assists IVs to have the capability to gather information from the environment and extract external features. The planning section helps IVs make more reasonable strategies from the starting point to the destination. The IVs can avoid obstacles and continuously optimize their driving behaviors to ensure safety and comfort for passengers. The control section makes the intelligent system the ability to control the vehicle smoothly. Perception, planning, and control are sequential and form the main body of AD technology. The computing system design acts as a crucial role in AD and IVs, which guarantees safety, security, energy saving, and communication efficiency. The communication module is a crucial tool for interaction with the outside world such as other vehicles, pedestrians, and smart transportation systems. The HD map is a guide for IVs, providing richer scene and location information just like the navigation application for human drivers. The testing module is an integral part of the process before IVs can operate on real roads. And human behaviors in IVs are a key part of current AD development, as understanding human behaviors can contribute to rational strategies for vehicles. 

\begin{figure}
\centering  
\includegraphics[width=9cm]{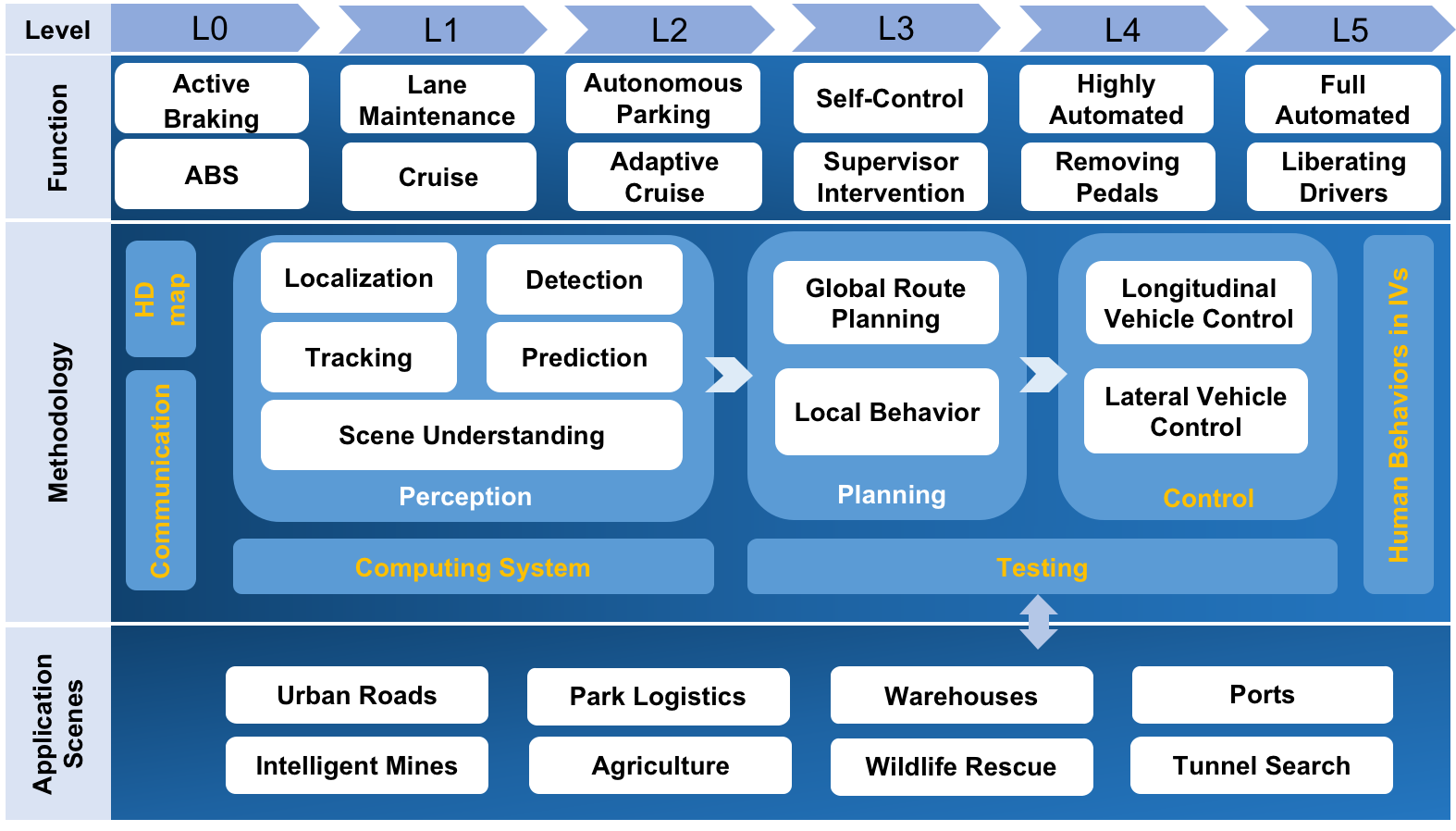}

\caption{The structure of autonomous driving with the function, methodology, and application scenes}
\label{fig:structure}
\end{figure}

The Society of Automotive Engineers (SAE) provides a taxonomy with detailed definitions of vehicle driving automation systems for six levels of driving automation, ranging from no driving automation (L0) to full driving automation (L5). We emphasize that the above partition method is based on the automated level, which is different from the definition of AD. Some researchers hold the idea that only motion platforms with L5 can be called AD vehicles. Some believe that L3 - L5 belongs to AD, while others hold with L1 - L5. We think that this difference may be due to the difference between academia and industry, and exists at the different development stages. However, there is no unified definition at present, and in this paper, we take the last view to collect articles, analyze characteristics and summarize methodology. Fig. \ref{fig:structure} shows the controlled mode and the main functions that can be achieved by the vehicle under different intelligence levels and a more detailed description could be found in Chapter VII human behaviors in intelligent vehicles.

In this paper, we consider $6$ sub-sections as independent chapters, and each of them includes task definition, functional divisions, novel ideas, and a detailed introduction to milestones of AD and IVs. The most important thing is that the research of them have rapidly developed for a decade and now entered a bottleneck period. We wish this article could be considered as a comprehensive summary for abecedarians and bring novel and diverse insights for researchers to make breakthroughs.

We summarize three contributions of this article:

\quad 1. We provide a more systematic, comprehensive, and novel survey of crucial technology development with milestones on AD and IVs.

\quad 2. We introduce a number of deployment details, testing methods, and unique insights throughout each technology section.

\quad 3. We conduct a systematic study that attempts to be a bridge between past and future on AD and IVs, and this article is the second part of our whole research (Part \uppercase\expandafter{\romannumeral1} for the survey).

%% file: body/methodology.tex



\input{body/methodology/control}

\input{body/methodology/system}

\input{body/methodology/communication}

\input{body/methodology/HDmap}

\input{body/methodology/testing}

\input{body/methodology/human-mechine}

%% file: body/methodology/control.tex
\section{Control}
Vehicle motion control is an important duty for enhanced driver assistance features. Many control tasks, such as lateral stability control and driving at the limits for accident avoidance, must be considered in the context of IVs. This section summarizes some of the major advancements in IV control during the previous few years. We will introduce the function of vehicle motion control, provide a comprehensive discussion on control strategies and discuss the state-of-the-art validation methods of control strategies.

\subsection{\textbf{Categories of Vehicle Motion Control}} 
We categorize vehicle motion control into two sub-parts, longitudinal vehicle control, and lateral vehicle control.

\subsubsection{Longitudinal Vehicle Control} Longitudinal control manages the acceleration of the vehicle through the vehicle’s throttle and brake to keep a safe distance behind another vehicle, maintain the desirable velocity on the road and apply the brake as quickly as possible to avoid rear-end collisions in emergencies \cite{survey_control_6}.


Longitudinal control strategies have been investigated in various scenarios, including platooning for connected IVs \cite{arefizadeh2018platooning}, speed harmonization \cite{malikopoulos2018optimal}, trajectory smoothing \cite{li2018piecewise}, and speed management on signalized arterials \cite{hao2018eco}. Local and string stability are crucial aspects of platoon longitudinal controllers. Local stability refers to a vehicle's capacity to maintain an equilibrium state under the condition of disturbance.  String stability refers to the magnitude of a disturbance that reduces or remains constant as it propagates through a platoon of vehicles. The control strategies for speed harmonization aim to determine the speed policy of the vehicles to prevent traffic breakdowns and mitigate the loss of highway performance. The fundamental idea of trajectory smoothing is to improve traffic flow stability and efficiency through the longitudinal control strategy of connected IVs. On signalized arterials, speed management systems aim to adjust signal timing to reduce stops at signalized crossings while smoothing vehicle trajectories. Common longitudinal control strategies include Fuzzy logic \cite{kim1996fuzzy}, Proportional–Integral–Derivative (PID) \cite{choi1996robust}, model predictive control (MPC) \cite{zhou2019distributed}, game theory \cite{huang2019game}, sliding mode control (SMC) \cite{wu2019path}, Fuzzy inference system (FIS) \cite{mar2001anfis}, Lyapunov-based adaptive control approaches \cite{swaroop2001direct}, and AI-approaches \cite{chen2017learning}. The challenges dealing with within the design of the longitudinal control system include string-stable operation with tiny headway, execution of longitudinal split, and join maneuvers subject to communication limitations \cite{dai2005approach, MotionPlanning}.


\subsubsection{Lateral Vehicle Control}
Lateral vehicle control systems focus on the vehicle’s position in the lane. The major function of the lateral vehicle control system is to keep the vehicle in the current lane (lane-keeping), drive the vehicle to the adjacent lane (lane change), or avoid collisions with its preceding vehicle \cite{gong2016constrained}. The control tasks for lateral control are to minimize the lateral displacement and the angular error with regard to the reference trajectory \cite{survey_planning_4}. Various linear and nonlinear controllers have been reported in the literature to control the motion of the vehicle including Game theory-based approaches \cite{zhang2019game}, PID \cite{marino2011nested}, MPC \cite{kim2014model}, $H_\infty$ robust control \cite{ganzelmeier2001robustness}, Lyapunov-based adaptive control approaches \cite{netto2004lateral}, Linear Parameter Varying (LPV) Controllers \cite{besselmann2009autonomous}, SMC \cite{liu2021super}, Fuzzy logic \cite{arifin2022steering} and deep learning-based approach \cite{eraqi2017end}.

As the lateral control system can change the vehicles’ lance safely and perform evasive maneuvers, it can offer potential benefits in reducing accidents and enhancing driving safety. However, full implementation of lateral vehicle control remains a real-world difficulty in lance changes and evasive maneuvers. For example, environmental factors such as weather conditions, road curves, and other various disturbances may affect the control performance, and a proper lateral control strategy is needed.

\input{tabels/methodology/control.tex}
\subsection{Control Methodology}
We introduce several common methodologies in vehicle control as Table \ref{table:control_table_1}.

\subsubsection{PID Control} 
The generic applicability of the PID controller can make it outperform other control approaches to in-vehicle control for IVs \cite{choi1996robust}. A nested PID steering control is developed for lateral control of IVs in the condition of unknown route curvature \cite{yu2018human}. PID control has the advantages of simple structures, easy implementation, and no need to know in-depth knowledge of the behavior of the system. However, PID parameters tuning is a challenge, and no optimal performance is guaranteed. In addition, it is hard for PID controllers to be adaptive and robust to unknown and changeable external environments \cite{huang2017parameterized}.

\subsubsection{Game-theoretic Approaches}
In game theory approaches the vehicles are regarded as game participants, and traffic rules are considered in the respective decision models. The decision models can mimic human behaviors by interacting with surrounding drivers, extract information from the interaction and generate the optimal control strategies, e.g. when and how to change the lane \cite{zhang2019game}. One advantage of the game-theoretical approaches is the consideration of the mutual influence of traffic participants, which can increase the reliability of decision-making. However, the dimension of the game can be highly increased by involving more participants, and the high dimensionality of the coupled game system may result in high computational complexity.

\subsubsection{Fuzzy Logic Control} 
Similar to PID controller, Fuzzy logic control doesn’t require the mathematical model of the plant allowing the controller to adequately deal with nonlinear vehicle dynamics. Another benefit is their humanoid control actions because of the human-like rules. Using membership functions, the input variables are translated into linguistic variables. The controller's output is determined by fuzzy rules that take the form of "if-then" statements. \cite{cai2009intelligent} presents a neuro-fuzzy controller with the aim to regulate the speed in order to maintain a safe distance with respect to the vehicle in front. \cite{sun2019nested} designs a fuzzy feedback controller to optimize the parameters of membership functions and rules in order to track the reference trajectories under different driving conditions. However, parameter optimization is challenging, and fuzzy inference rules with a lot of expert knowledge are also required for fuzzy logic control \cite{guo2012design}.

\subsubsection{MPC}
MPC is extremely popular for vehicle control, see longitudinal control \cite{ arefizadeh2018platooning}, lateral control \cite{yu2018human}, and integrated control \cite{8885839}. The principle of MPC is to find a predictive motion solution over a longer horizon period by solving the problem at each sample time and applying the first sequence of actions. In this way, MPC simulates a receding horizon control and changes the solution set to remain accurate to upcoming information \cite{nobe2001overview}. \cite{kim2014model} proposes an MPC-based path-tracking controller that incorporates the dynamic characteristics of the steering actuation system to ensure accurate and smooth tracking. The main advantage of MPC algorithms is their ability to consider multiple performance criteria of control efficiency, ride comfort, fuel consumption, and their constraint handling capability on vehicles’ physical limits and safety.  However, they are still inadequate for non-convex and high-complexity issues. Excessive computing difficulties may impede their real-time applications.

\subsubsection{SMC Approaches}
SMC offers the capacity to adapt to unknown disturbances and matched uncertainties. A series of path-following controllers based on SMC is proposed in \cite{wu2019path} and simulation results show that the SMC controllers are robust to disturbances and provide proper path-tracking performance.  However, the model uncertainties and the disturbances are time-varying in the realistic scenario and the model accuracy greatly influences the control performance. The control performance may be degraded due to the modeling errors \cite{rausch2017learning}.

\subsubsection{LPV Control}
The LPV controller is a linear controller and has been designed together with predictive control for lateral control \cite{ganzelmeier2001robustness} and integrated system control \cite{gaspar2005design}. This method requires advanced knowledge of a plant model as well as real-time signals from all states during driving. A state observer is required because this is not provided by default.

\subsubsection{AI-guided Driving Policy Learning}
Research on artificial intelligence algorithms is progressing rapidly because of the commonplace of big data. Compared to traditional control systems, AI-guided vehicle systems do not require any system-level knowledge and have shown great potential to optimize nonlinear systems in complex and dynamic environments \cite{chen2017learning,Hierarchical}. Instead of designing explicit hand-engineered algorithms for lane detection, obstacle detection, path planning, and control law design separately, the AI-guided vehicle systems can combine all efforts within one framework and teach the system to perform well and generalize to new environments through learning \cite{nobe2001overview}. \cite{yu1995road} investigates a road-following system using reinforcement learning. The testing results show that the proposed algorithm can use previous driving experiences to learn how to drive in new environments and improve road-following performance through online learning. \cite{wang2018reinforcement} uses Deep Q-Network (DQN) to train a vehicle agent to learn automated lane change behaviors. \cite{chae2017autonomous} presents a novel autonomous braking system that uses deep reinforcement learning to intelligently manage the vehicle's velocity when a potential collision is announced. \cite{naranjo2006acc+} investigates an end-to-end controller based on a convolutional neural network (CNN) to provide steering commands without engineering each autopilot component. The AI-guided control strategies are very powerful and their capacity to generalize over a variety of similar scenarios makes them fit for minor deviations in the driving conditions \cite{samak2011control}. However, safety validation of these AI-guided systems is very challenging due to the opaque nature of the methods and there is no explanation for failure can be given. In addition, the need for training data may hinder the research of AI-guided systems.

The classic PID and sliding mode control method may fail to adapt to complex environments and unknown disturbances. The MPC and LPV are based on precise mathematical models, the uncertain and nonlinear mathematical model of the vehicle during real-world driving may degrade the control performance, especially in extreme maneuverings. The fuzzy rules of the fuzzy logic systems are based on experience, and there are no qualitative rules to refer to, making them challenging to use in engineering applications. Due to the high dimensionality of the coupled game system, the design of game theory-based control strategies is challenging. The AI-based method requires a significant amount of off-line computation and training and more study into interpretability and functional safety validation methodologies for AI-driven vehicles is needed.

%% file: tabels/methodology/control.tex
\begin{table}[]
\caption{Control Methodology with advantages and disadvantages}

\begin{tabular}{cll}
\hline
Method        & Advantages                                                                                                            & Disadvantages                                                                                       \\\hline
PID & \begin{tabular}[c]{@{}l@{}}1.Simple structures\\2.Easy implementation\\3.Non-mathematical mode\end{tabular}           &\begin{tabular}[c]{@{}l@{}}1.Parameters tuning \\2.Non-optimal performance \\3.Weak robustness\end{tabular}                                    \\\hline
Game & \begin{tabular}[c]{@{}l@{}}1.Learning by imitation\\2.Optimal strategies\end{tabular}                               & High complexity                                                                       \\\hline
Fuzzy    & \begin{tabular}[c]{@{}l@{}}1.Non-mathematical mode. \\2.Humanoid control actions\end{tabular}                 & \begin{tabular}[c]{@{}l@{}}1.Parameter optimization\\2.Prior knowledge\end{tabular}  \\\hline
MPC            & \begin{tabular}[c]{@{}l@{}}1.Control efficiency\\2.Multiple factors\end{tabular}                                    & \begin{tabular}[c]{@{}l@{}}1.Inadequate for non-convex \\2.High complexity\end{tabular}           \\\hline
SMC            & \begin{tabular}[c]{@{}l@{}}1.Robustness\\2.Suitable for unknown space\end{tabular}                                  & Heavy dependence on models                                                                          \\\hline
LPV            & 1.Linear controller                                                                                               & Advanced knowledge                                                                     \\\hline
AI      & \begin{tabular}[c]{@{}l@{}}1.No system knowledge \\2.No manual setting\\3.Strong adaptability\end{tabular} & \begin{tabular}[c]{@{}l@{}}1.Hard to validate\\2.Poor interpretation\end{tabular}   \\\hline
\end{tabular}         
\label{table:control_table_1}
\end{table}

%% file: body/methodology/system.tex
\section{Computing System Design}
In this section, the computing system design of the IV will be described from three essential aspects of IVs, namely, the computing systems architecture, and sensor systems with various applications.

\subsection{Computing System Architecture}
\begin{figure}
\centering  
\includegraphics[width=8.5cm]{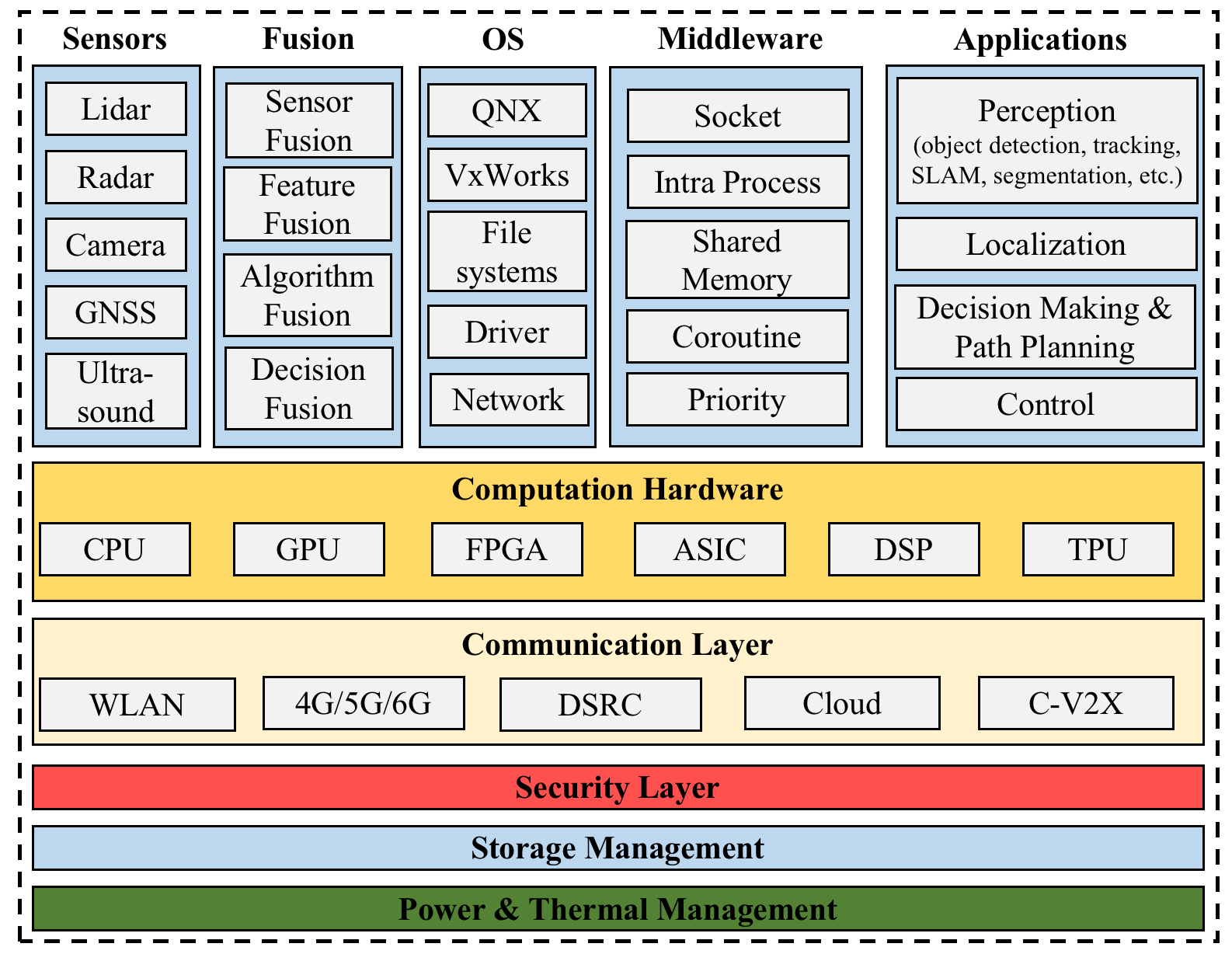}

\caption{Typical computing architecture for IVs \cite{survey_8, survey_system_1, survey_system_2}}
\label{fig:system_fig_1}
\end{figure}

The computing system plays a vital role in IVs to guarantee safety, security, energy, and communication efficiency. For a typical IV which usually can be equipped with a large number of onboard sensors, including Lidar, cameras, Radar, communication module, and Global Navigation Satellite System (GNSS), etc., the data generated per minute can be huge. To ensure the efficient processing and fusing of heterogeneous information, the computing system must be well-designed to process the information in real time and maintain safe AD. According to \cite{survey_8}, there are two basic types of computing architecture that are widely used in IVs, which are the modular design method and the end-to-end design method. The modular design approach decouples the functional units into separate modules which can be easier for system implementation, fault diagnosis, and module update. Based on the modular design approach, the computing architecture of IV can be separated into the following key modules (as shown in Fig. \ref{fig:system_fig_1}), including computation, communication, storage, security and privacy, and power management. While the end-to-end computing architecture is largely motivated by current artificial intelligence and mainly relay on learning-based approaches to process the sensing data and generate control output directly \cite{system_2}. Considering the modular design approach is much more widely used and mature for computing architecture design on IVs, the following concepts of computing systems are reviewed based on this branch.

\subsubsection{Computing Hardware}
There is multiple powerful computing hardware to support the real-time computation capability of IVs. Typically, the Central Processing Unit (CPU), Graph Processing Unit (GPU), Field Programmable Gate Array (FPGA), Digital Signal Processor (DSP), Application Specific Processor Unit (ASIC), and its custom-developed deep learning-oriented Tensor Processing Unit (TPU) are the widely used computing hardware on IVs nowadays \cite{survey_system_2}. Specifically, Nvidia published the Nvidia Drive series to support IV technology with powerful GPU support \cite{system_9}. Xilinx released the Zynq UltraScale+MPSoC automotive-grade processor based on FPGA, which achieved more energy-saving performance (14 images/W) than Nvidia Tesla K40 GPU (4 images/W) \cite{system_10}. The Mobile EyeQ6 series and TPUv4 are the leading ASIC-based solution for IVs, which show significant improvement in the computation power and energy efficiency \cite{system_11}. Huawei published its Mobile Data Centre (MDC) solution for real-time computing on L4 IVs. The MDC is equipped with a Huawei Ascend AI chipset which is capable of up to 352 TOPS of computing power, and industry-leading 1 TOPS/W energy efficiency. Horizon Robotics also published their cutting-edge automotive system-on-a-chip (SoC) processor, Horizon Robotics Journey 5, which is a particular design for optimizing the runtime execution of onboard vision and Lidar-based perception tasks. The HRJ-5 features up to 128 TOPS processing capabilities for a single processor and supports neural computing from 16 sensors.

\subsubsection{Operating System}
Real-time operating system (RTOS) is a key module for the computing systems on IVs that enables resource management and real-time perception, planning, decision-making, and control autonomously at a fast speed \cite{survey_system_2, OS}. OS on IVs should support real-time resource allocation, networking, filing, communication, etc. Two common RTOS kernels are QNX and VxWorks \cite{system_12}. The QNX is widely used in the automotive industry that contains CPU scheduling, interprocess communication, interrupt redirection, and timer \cite{system_12}. VxWorks is a monolithic kernel and is designed for real-time embedded systems and also can be used for safety and security \cite{system_13}. It is designed with a shared memory architecture and it supports multiple architectures like Intel, POWER, and ARM. 

\subsubsection{Middleware}
In the middle of RTOS and application layers, the middleware layer is needed to bind the multiple AD services together to manage the application communication and schedule the sensing and computing resources. In \cite{system_7}, three influential middleware architectures, namely Robot Operating System (ROS1), ROS2, and Cyber are evaluated regarding the communication latency for AD applications. The Inter-Process Communication mechanism in ROS1 provides high compatibility and extensibility \cite{system_15}. ROS2 shows efficient performance in real-time distributed systems based on the Data Distribution Service (DDS)-based communication scheme \cite{system_16}. Then, the Cyber is a recently published middleware architecture by Baidu Apollo, which is primarily designed for AD. 

\cite{system_7} summarised the three architectures based on a comparison study, it was found that 1) the communication through a shared memory mechanism is the optimal solution considering the performance and reliability of IVs; 2) Data copy and serialization/deserialization operation in the middleware layer are the main attributes of communication overhead. 3) Optimization in communication characteristics can bring significant benefits to computation efficiency. 4) the higher the performance of the platform, the more likely the communication becomes the bottleneck for processing speed. 5) Additional latency can be introduced if the publishing frequency and corresponding processing time are not well calibrated. 

In sum, middleware can be a contributor to the end-to-end latency and to avoid becoming the performance bottleneck for the overall computing system, the communication and management in the middleware should be carefully designed.

\subsubsection{Metrics for Computing Systems on IVs}
In \cite{survey_system_1}, the authors identified seven evaluation metrics for the computing system on IVs, which are accuracy, timelines, power, cost, reliability, privacy, and security. A brief summary of the characteristics of these metrics is demonstrated in Table \ref{table:system_table_1}. 

\input{tabels/methodology/system_tabel_1}

\subsubsection{Constraints for Computing Systems on IVs}

Though computing systems on IVs have achieved significant improvement in recent years, there are still several essential constraints that exist in the hardware, software, and sensor layers, etc. which can delay the wide deployment of commercial IVs. For example, In \cite{survey_system_1}, five major design constraints of the computing system, namely, performance constraints, predictability constraints, storage constraints, thermal constraints, and power constraints for IVs are identified and analyzed. Therefore, in this section, we will jointly summarize and analyze these challenges that exist in the different layers of the computing system.

Specifically, performance constraints mainly exist at the application level as shown in Fig. \ref{fig:system_fig_1}. In terms of the fundamental functions of IVs, including perception, planning and decision-making, and control, the gap between machine and human performance still exists. It is stated that the frame rate and processing latency are the two main performance constraints for the IVs. The human driver can react to an event with the action within 100-150ms, therefore, for safety concerns, the vehicle should react even faster than the human driver within a latency of 100ms at a frequency of at least once every 100ms as well \cite{survey_system_1}. 

Predictability is another constraint that is defined from both the temporal aspects (meeting the time deadline) and functional aspects (making the correct decision). Despite the conventional mean latency metric, the tail latency (99.99th-percentile latency) should also be used to satisfy the stringent predictability requirement \cite{survey_system_1}.

The storage constraint is an essential aspect and bottleneck for energy-saving and computing performance on IVs. It is shown that one IV can generate 2 to 40 TB of data per day \cite{survey_8}. The tens of TBs of storage space per day would require high storage speed and space for the vehicle. Besides, real-time data storage and transfer would also require a significant boost in energy and power usage. 

The thermal constraints of the IVs can be a serious consideration as well. There are two typical aspects of the thermal constraints which are 1) the temperature to keep the computing system working at operational range, and 2) the heat generated by the computing system should lead to limited impact on the vehicle thermal profile \cite{survey_system_1}. It is shown that without the cooling system, every 1 kW power consumption by the computing system would increase the in-cabin temperature up to 10℃ per minute \cite{system_17}. Therefore, usually, the computing system should be implemented in a climate-controlled area which means an additional cooling system is needed to mitigate the thermal impact of the computing system \cite{ system_17}. 

Last, power constraint is another critical aspect that can significantly influence the capability of IV. According to \cite{survey_system_1}, the power constraint of the computing system contains three major aspects which are power consumption of the computing system, consumption of the storage, and cooling overhead. The computing consumption and additional energy consumption for storage and cooling would dramatically decrease the millage of IVs, especially the electric IVs. For instance, the miles per gallon (MPG) rate can be reduced by one for every additional 400 W energy usage, and an energy-hungry system like GPUs can reduce the fuel efficiency by up to 11.5\% \cite{survey_system_1}. Therefore, more energy-efficient computing systems and green AI techniques are extremely expected for IVs in the future. 

\subsection{Sensor Systems}
Sensor systems on IVs provide essential sensing and detection data to the computation units. Common onboard sensors for IVs include Lidar, Radar, cameras, ultrasonic, inertial measurement units (IMU), GNSS, etc. Dues to the large uncertainty properties during driving, IVs should rely on multiple sensor fusion to guarantee driving safety. Hence, in this part, the characteristics of common sensors are reviewed.

\subsubsection{Sensors in IVs}
De Jong et al. classified the sensors on IVs into smart sensors and non-smart sensors according to the internet-of-things concept in \cite{survey_hardware_3}. Smart sensors on IVs refer to those that can provide extra detection or perception information, such as object detection/tracking, and other event description from the camera, Lidar, or Radar systems. On the contrary, non-smart sensors are those which only provide raw data for processing. The study of characteristics of onboard sensor systems has been widely studied in \cite{survey_hardware_3}. The typical sensor fusion methods are shown in Fig. \ref{fig:system_fig_1}, and can refer to \cite{survey_hardware_3, survey_dynamic_15} for more detailed discussion. A brief description of the different sensors and selected review studies are shown in Table \ref{table:system_table_2}. 

\input{tabels/methodology/system_tabel_2}

\subsubsection{Computing System Design for IVs with Various Application Scenarios}
The computing system design of IVs is also heavily dependent on the application scenarios, policies, and regulations. Nowadays, the most commonly used standard for IVs regarding safety, communication, and design can be found in \cite{system_47}. The standard roadmap in \cite{system_47} covers the current and ongoing standards for safety and assurance, perception, decision-making, data, security, infrastructures, and human factors. It should also be noticed that the design of the IV shall consider the specific application domains as well. For example, the autonomous mining truck and its testing techniques have been summarized in \cite{system_48}, which shows significantly different considerations compared to passenger cars. Other successful applications of IVs at airports, harbors, and for logistic purposes can be found in \cite{ system_51}. Clearly, each specific IV requires significantly different concerns in the overall computing system design and should respect the specific real-world application environment. 

Another emerging research areas for sensor systems in IVs are the design of the explainable interface for IVs. On-board explanation of the perception, localization, decision-making, and control based on the multi-modal signals from the sensor systems is an essential design requirement for future IVs to enhance the human trust and acceptance of the IVs for worldwide commercialization \cite{survey_11}. The Explainability of IVs can be further divided into interpretability and completeness. Further, interpretability has two main branches, namely, transparency and post-hoc explanation \cite{survey_explainable}. Meanwhile, local explanation and global explanation also requires different computing system design for the IVs \cite{survey_explainable}. In sum, the design criteria for an explainable interface of IVs will contribute to the societal and legal requirements of the IVs, and more importantly enable safer, transparent, public-approved, and environmentally friendly IVs \cite{system_54}.

%% file: tabels/methodology/system_tabel_1.tex
\begin{table}[]
\caption{Metrics for Computing Systems for IVs \cite{survey_8}}
\begin{tabular}{cm{6.7cm}}\toprule

Metric      & \multicolumn{1}{c}{Description}                                                                                                             \\ \hline
Accuracy    & Usually evaluate the difference between the predicted value and ground truth in the precision, recall, and other metrics.                                                                 \\ \hline
Timelines   & Evaluate the inference and processing speed of the computing system to ensure the computation can be finished before the deadline and minimize latency and tail latency.                     \\\hline
Power       & Evaluate the power usage of the computing system and analysis its impact on the energy, mileage, and comfort issues for IVs.                                                                      \\\hline
Cost        & Mainly refer to the financial cost for the computing system toward board deployment.                                                                                                              \\\hline
Reliability & To guarantee the safety of the IVs, the worst-case execution time, interruption or emergency stop capabilities, and fault-tolerant should be considered.                                          \\\hline
Privacy     & To protect the privacy of all the road users including both in-cabin and surrounding users' private data. Also, the acquisition, storage, and communication privacy should be further considered. \\\hline
Security    & The onboard security can be mainly divided into four parts: sensing security, communication security, data security, and control security.        
\\ \hline
\end{tabular}
\label{table:system_table_1}
\end{table}

%% file: tabels/methodology/system_tabel_2.tex
\begin{table*}[]
\caption{Comparative Study for Common Sensor Systems on IVs.}
\begin{tabular}{cp{5cm}p{5cm}p{5cm}}\toprule
Sensors    & \multicolumn{1}{c}{Advantages}                                                                                                                                                                     & \multicolumn{1}{c}{Disadvantages}                                                                                                                                                                           & \multicolumn{1}{c}{Review Studies}                                                                                                           \\ \hline
Lidar      & \begin{tabular}[l]{@{}l@{}}1.Support high-resolution 3D point cloud\\ 2.Larger and wider sensing range\\ 3.Accurate object and distance detection\\ 4.Robust to lightning\end{tabular} & \begin{tabular}[p{5cm}]{@{}l@{}}1.High Cost\\ 2.Poor performance to harsh weather\\ 3.A large amount of data stream\end{tabular}                                      & \begin{tabular}[p{5cm}]{@{}l@{}}1.Point Cloud and networks \cite{survey_dynamic_16}\\ 2.SLAM \cite{system_27}\\ 3.Sensor design \cite{system_24}\end{tabular} \\ \hline

Camera     & \begin{tabular}[p{5cm}]{@{}l@{}}1.Cost-effective\\ 2.Visual recognition and texture extraction\end{tabular}                                             & \begin{tabular}[p{5cm}]{@{}l@{}}1.Easily affected by lighting and weather.\end{tabular}                                                       & \begin{tabular}[p{5cm}]{@{}l@{}}1.Perception and networks \cite{survey_dynamic_21}\\ 2.Fusion \cite{survey_dynamic_15}\end{tabular}                         \\ \hline

Radar      & \begin{tabular}[p{5cm}]{@{}l@{}}1.Support extreme long sensing range\\ 2.Resilient to various external conditions\end{tabular}                                          & \begin{tabular}[p{5cm}]{@{}l@{}}1.Smaller data size \\ 2.Indistinguishable from static targets\end{tabular} & \begin{tabular}[p{5cm}]{@{}l@{}}1.Detection and perception \cite{survey_hardware_2}\\ 2.Radar Communication \cite{system_35, system_36, system_37}\end{tabular}           \\ \hline

GNSS        & \begin{tabular}[p{5cm}]{@{}l@{}}1.Low cost and stable\\ 2.Non-accumulation error for GNSS\end{tabular}                                                                                   & \begin{tabular}[p{5cm}]{@{}l@{}}1.Relative low accuracy for GNSS and IMU\\ 2.Slow update frequency \\3. Requiring an unobstructed view\end{tabular}                                      & \begin{tabular}[p{5cm}]{@{}c@{}}1.Localization \& navigation \cite{system_38, system_39}\end{tabular}                                                                   \\ \hline

Ultrasound & \begin{tabular}[p{5cm}]{@{}l@{}}1.Low cost\\ 2.High accuracy for close range\\ 3.Resilient to adverse weather conditions\end{tabular}            & \begin{tabular}[p{5cm}]{@{}l@{}}1.Short-range detection\\ 2.Inapplicable for high speed\end{tabular}  & \begin{tabular}[p{5cm}]{@{}l@{}}1.A systematic review \cite{system_40}\\ 2.All-weather conditions \cite{system_41}\end{tabular} 
\\ \hline
\end{tabular}
\label{table:system_table_2}
\end{table*}

%% file: body/methodology/communication.tex
\section{Communication}
Facing the complex traffic environment, the perception of an individual IV is blinded by obstacles and severe weather, which affects the safety of AD. On the other hand, the decision-making capability of an individual vehicle is limited by the onboard computing and storage resources, which makes it difficult to cope with the challenges of the large density and mixed traffic flow environment. Therefore, IVs not only need to optimize their intelligence level but also need to expand their sensing and decision-making capabilities by obtaining external information and resources through wireless communication technologies. Different from the mobile terminals in traditional cellular networks, the high-speed movement of vehicles leads to rapid changes in network topology and frequent switching of communication links. Meanwhile, the complex and variable driving environment of vehicles leads to the multipath effect and interference from other signals for vehicular communication. In addition, safety-related applications such as collision avoidance and platooning of IVs require networks to meet the stringent requirements of low latency and ultra-high reliability, which are difficult to achieve with traditional wireless networks. Therefore, it is necessary for IVs to adopt dedicated communication technologies to ensure efficient and stable interactions between vehicles and other vehicles, infrastructure, and cloud platforms.

As a powerful extension of the perception capabilities of AD and IVs, the communication of IVs not only involves the vehicles themselves, but also involves a variety of infrastructure and technical elements of transportation, communication, and other systems, and is a focused area for the integration and convergence of automotive, transportation, communication, information, and other industries. Vehicular communication technology originated from academic research and demonstration projects conducted in Europe and the United States at the end of the 20th century that focused on vehicle-to-vehicle (V2V) and vehicle-to-infrastructure (V2I) communications, which was known as vehicular ad-hoc network (VANET) \cite{hartenstein2008tutorial} or vehicular network at that time. With the development of novel information and communication technologies, the connection and cooperation of all elements in transportation systems have been further strengthened to form a connected and cooperative AD, the concept of VANET was extended to the Internet of vehicles and it has gained wide attention due to its supporting role for industries such as intelligent transportation and AD.

\begin{figure}
\centering  
\includegraphics[width=8.5cm]{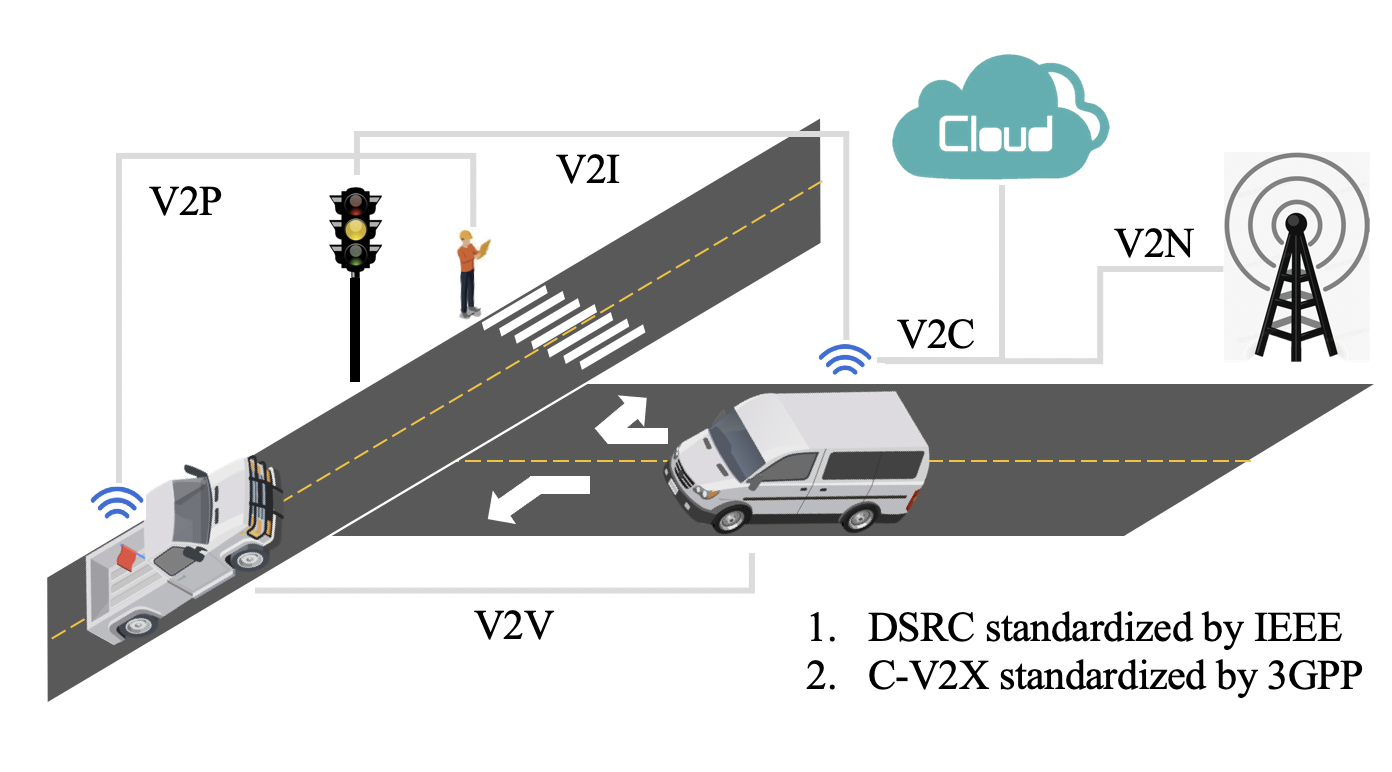}

\caption{Elements of vehicle-to-everything (V2X) for vehicular communication}
\label{fig:communication}
\end{figure}

With the expansion of the concept, the vehicular communication technology has evolved from the VANET that connects vehicles and infrastructure to the vehicle-to-everything (V2X) \cite{abboud2016interworking} that connects various elements of the transportation system such as vehicles, infrastructures, pedestrians and clouds. Specifically, V2X contains two types of technologies, dedicated short-range communication (DSRC) standardized by IEEE and cellular V2X (C-V2X) standardized by the 3rd Generation Partnership Project (3GPP) \cite{chen2020vision}. In terms of communication methods, V2X communication can be subdivided into V2V, V2I, vehicle-to-pedestrian (V2P), and vehicle-to-cloud/network (V2C/N) communication as shown in Fig \ref{fig:communication}. Therefore, vehicular networks can realize real-time and efficient information interaction among pedestrians, vehicles, infrastructures, and clouds, and support the massive data transmission of AD and IVs.

\subsection{Standardization of V2X Communication}
To achieve efficient communication and interaction between heterogeneous vehicles, it is necessary to establish unified protocols and standards covering the physical layer to the application layer. As mentioned above, based on different radio access technologies, there are two major types of V2X communication standards, namely DSRC and C-V2X. In this section, we briefly present the status and development of these two types of standards.


\subsubsection{DSRC}
Strictly speaking, DSRC is a radio access technology for vehicular communication, but it is also used to broadly refer to vehicular communication based on this radio access technology. Currently, the technology is relatively mature, and several organizations for standardization, such as IEEE, Society of Automotive Engineers (SAE), and European Telecommunications Standards Institute (ETSI), have worked on DSRC-related standards \cite{abboud2016interworking}. Here we take the standard of the United States as an example to describe its standardization. In the United States, IEEE has developed the IEEE 802.11p and IEEE 1609 standards for the 5.9 GHz spectrum and formed the WAVE protocol stack \cite{uzcategui2009wave}. IEEE 802.11p is a communication protocol expanded from the IEEE 802.11 standard to support vehicular communication in a mobility environment \cite{survey_communication_7}. The IEEE 1609 protocol family is a high-level protocol for WAVE, including 1609.1, 1609.2, 1609.3, 1609.4, etc. \cite{teixeira2014vehicular,dibaei2022investigating}. Specifically, 1609.1 is a standard on resource management at the application layer of WAVE, 1609.2 defines secure message formats and their processing, 1609.3 defines routing and transport services, and 1609.4 mainly provides a communication standard for multi-channel cooperation. It is worth noting that the Federal Communications Commission (FCC) has already reallocated the 70 MHz dedicated spectrum for DSRC in 2019 and dedicated 20 MHz of that band for C-V2X technology. In 2020, the FCC added another 10 MHz to the previous proposal, formally allocating 30 MHz of spectrum to C-V2X communications and dropping support for DSRC communications \cite{fcc2020use}.

\subsubsection{C-V2X}
The standardization of C-V2X technology by 3GPP can be divided into two phases, LTE-V2X and NR-V2X, which complement each other and are designed with forwarding and backward compatibility considerations \cite{gyawali2021challenges}. C-V2X mainly includes two types of communication methods, cellular mobile communication (using Uu interface) and sidelink communication (using PC5 interface) \cite{hasan2020securing}.
The standardization of LTE-V2X was initially completed in 3GPP Release 14, and its main application scenario was safety-oriented services for intelligent driving assistance. Release 15 enhanced the LTE-V2X technology and supported some enhanced applications of vehicular communication. To support the requirements of advanced V2X services and AD, 3GPP Release 16 and 17 carried out research on NR-V2X and enhanced the PC5 and Uu interfaces based on 5G NR, and the related standards are frozen now \cite{chen2020vision}.
LTE-V2X, based on 4G cellular communication, is the first C-V2X communication technology, which defines the basic architecture and technical principle of C-V2X combining cellular mobile communication and sidelink communication methods \cite{chen2016lte}.
NR-V2X, based on 5G communication, is an evolutionary version of LTE-V2X. It follows the system architecture and key technical principles of combining the Uu and PC5 interfaces defined in LTE-V2X. Meanwhile, to support a series of enhanced V2X services such as platooning, extended sensors, and remote driving for the future, NR-V2X is enhanced with novel technologies of 5G NR on the PC5 interface to provide higher data rate, low latency, and ultra reliable communication services. In terms of key technologies, NR-V2X supports unicast and multicast communication modes in addition to broadcast, and adaptive improvements have been made to resource allocation and synchronization mechanisms \cite{naik2019ieee}.

\subsection{Innovative Studies of Vehicular Communication}
Facing the stringent requirements of AD and IVs for vehicular communication, researchers have proposed improvements and optimizations for the existing vehicular communication methods in several research directions of communication and networks \cite{ahmed2018cooperative}. In this part, we briefly present innovative studies for vehicular communication in terms of the physical layer, media access control, routing, and security.

\subsubsection{Physical Layer}
The physical layer is one of the important parts that affect the performance of vehicular communication, which enables the transmission of vehicle and traffic data through radio channels \cite{liang2017vehicular}. Currently, orthogonal frequency division multiplexing (OFDM) is the major modulation technology in vehicular communication \cite{wang2019networking}, thus many studies are devoted to optimizing OFDM parameters to ensure reliable and efficient data transmission \cite{arslan2017effects}. In addition, facing the requirement of data rate for communication in AD and IVs, millimeter-wave communication \cite{ghafoor2020millimeter} and visible light communication (VLC) \cite{memedi2021vehicular} in the vehicular environment have become emerging research directions.

\subsubsection{Congestion Control}
Since the bandwidth of vehicular communication is limited, with large-scale deployment of vehicles, the access of massive terminals can lead to channel congestion and cause delay or even failure of data transmission \cite{paranjothi2020survey}. Congestion control avoids the congestion in the channel and improves the stability and reliability of data transmission through power-based, rate-based, hybrid, priority-based, and cross-layer approaches \cite{rashmi2021survey}.

\subsubsection{Resource Allocation}
Facing the dynamic channel access and multi-user interference caused by the high mobility of massive vehicles, vehicular communication maximizes the system performance by optimizing the allocation of resources such as channels, bandwidth, and transmit power \cite{ahmed2018cooperative}. For DSRC, various studies have proposed schemes such as medium access control (MAC) parameter allocation, channel allocation, and rate allocation \cite{noor2022survey}. For C-V2X, resources are efficiently utilized through centralized schemes based on clustering, cloud computing, non-orthogonal multiple access (NOMA) and semi-persistent scheduling, and distributed schemes based on location, sensing, and enhanced randomization \cite{le2021comprehensive}. In addition, machine learning, especially deep reinforcement learning, as a powerful analytical technique for complex problems, has become an important scheme for resource allocation and optimization \cite{nguyen2021drl}.

\subsubsection{Routing}
In vehicular communication, the range of one-hop communication between vehicles and other units (vehicles or infrastructure) is limited, so routing is needed to achieve long-distance data interaction between vehicles and vehicles or roadside infrastructures \cite{jeong2021comprehensive}. However, facing the frequently changing topology and unstable connections caused by the high-speed movement of vehicles, the traditional routing methods in wireless communication are no longer applicable \cite{xia2021comprehensive}. Therefore, many studies have proposed enhanced routing schemes in vehicular communication to achieve efficient and stable data exchange, which include position-based, topology-based, geocast-based, broadcast-based, and cluster-based routing protocols \cite{sun2021applications}. Various routing protocols improve the quality of service (QoS) of vehicular communication by optimizing the delay, distance, reliability, energy consumption, or security of routing \cite{belamri2021survey}. In addition, with the development of machine learning, learning-based routing protocols have also attracted more attention \cite{kayarga2021study}.


\subsubsection{Security}
IVs interact with traffic systems through vehicular communication to improve driving safety and traffic efficiency \cite{hasan2020securing}. In this environment, attacks such as replay attacks, man-in-the-middle attacks, impersonation attacks, spoofing attacks, and malicious third-party attacks threaten the information security and even driving safety of vehicles \cite{gyawali2021challenges}. Therefore, facing the security requirements of vehicular communication, various studies prevent malicious attacks in terms of authentication, authorization, confidentiality, data integrity, and availability \cite{gyawali2021challenges}. In addition, machine learning-based attack detection and blockchain-based consensus and tamper-proof have become emerging research directions \cite{talpur2022machine}.

%% file: body/methodology/HDmap.tex
\section{HD map}

HD Maps are an integral part of modern AD. The information from the map supports several functions of AD, including localization, perception, planning, control, and systems management. Different from the navigation maps from commercial applications, HD maps serve the AD system instead of human drivers. Besides, HD maps could achieve a high level of precision within a centimeter involving multiple elements in the road such as street signs, signal lights, bridges, road guardrails, trees, corners, motion objects, etc.

\subsubsection{Composition and Generation process}

The production steps of HD maps can be divided into four main loops, data acquisition, map generation, semi-automated correction, and validation. After the sensors (Lidar, camera, radar, GNSS, odometry, etc.) installment, calibration, and synchronization, the collection platform could drive around the target region and store the sensor data. Then the clear, concatenation method, detection, and segmentation algorithms are utilized to generate a coarse base map. Introducing semi-automated correction and artificial verification, lane traffic signs, and logical information could be added to the map. Through a series of testing and validation processes, the HD map could be released and employed in the AD system on IVs. And the HD map could be updated and removed when road elements and the region are changed.

\subsubsection{Types and Standards}
According to the update rate of the elements, the HD map can be divided into five layers: base map, geometric map, semantic map, map priors, and real-time. The geometric map is composed of raw sensor data on the base map (initial map). The semantic map is constructed upon the former layer by introducing static semantic information such as lane boundaries, intersections, parking spots, stop signs, traffic lights, etc. The map priors layer contains dynamic information and human behavior information such as the change order of traffic lights, the average wait times, the probability of a vehicle at a parking spot, the average speeds of vehicles at parking spots, etc. Autonomy algorithms commonly consume these priors in models as inputs or features and are combined them with other real-time information. The real-time knowledge layer is the topmost layer in the map that is dynamically updated with real-time traffic information.

Geographic Data File (GDF) has been released by ISO/TC 204 which provides a basic version for the storage and exchange of map information. GDF-5.1 obey the Local Dynamic Map (LDM) standard and involves a number of information such as the weather, traffic conditions, static elements, etc. Open AutoDrive Forum (OADF) actively promotes the standardization of HD maps as a cross-domain platform. Traveller Information Services Association (TISA) attempts to increase HD map accuracy by introducing the Transport Protocol Experts Group (TPEGTM). Advanced Driver Assistance Systems Interface Specification (ADASIS) Forum releases the V3 protocol to support the distribution function within IVs. In addition, Navigation Data Standard (NDS) and OpenDRIVE are two major industrial standards for the generation of HD maps.

\subsubsection{Data quality}

There are 5 measuring standards to evaluate the quality of AD maps: 1) accuracy, 2) precision, 3) completeness, 4) consistency, and 5) timeliness. Accuracy is the deviation between actual and mapped values. Precision means the smallest discernible unit of a map. Completeness refers to whether the map contains all the real-world features. Consistency indicts the logical rules of data structure, attributes, and relationships. Timeliness means the difference in time between construction and use. In order to increase data quality, Wong and Ellul \cite{HDmap_81} proposed geometry-based metrics as part of the fitness assessment for 3D maps. Javanmardi et al. \cite{HDmap_79} defined four criteria to evaluate the vehicle location capability of maps, from the view of the layout, feature adequacy, presentation quality, and local similarity. Murphy and Pao \cite{HDmap_82} proposed a method for detecting unmapped or mismapped roads and parking lots in the context of map matching, resulting in a system that is more robust and can even correct errors in the underlying map. From a similar but alternative perspective, Yang and Huang \cite{HDmap_80} studied how to make IV systems more resistant to malicious attacks on their sensor.

%% file: body/methodology/testing.tex
\section{Testing}

\begin{figure}
\centering  
\includegraphics[width=8.5cm]{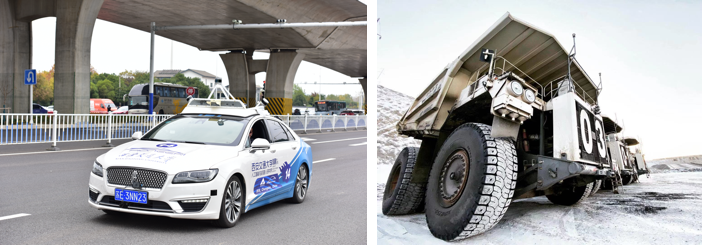}

\caption{The figure shows the appearance of IVs. The left vehicle is the test autonomous platform of Xi'an Jiaotong University. The right vehicle is the autonomous mining truck at an open-pit mine in China.}
\label{fig:system_fig_0}
\end{figure}

Vehicle testing has become a popular research topic in the field of IVs since 2016. On one hand, several companies that produce IVs encountered vehicle accidents during testing or running, which rises urgent demand to identify causes of accidents and fix complicated IV systems. On the other hand, researchers are increasingly interested in intelligence testing of generalized intelligent systems.  (Artificial Intelligence Test A Case Study of IVs; Parallel Testing of Vehicle Intelligence via Virtual-Real Interaction).

\subsection{Testing Platforms}
To verify vehicle algorithms, especially for the motion control algorithms, a series of testing should be designed for control strategy validation. In terms of testing patterns, the validation type can be divided into simulation studies and experimental tests. Testing in the actual world can be costly in terms of time, labor, and money. Simulation studies, in contrast, are much less expensive, faster, more adaptable, and can be used to create scenarios that are difficult to replicate in real life \cite{wang2018reinforcement}.

\subsubsection{Simulation Platforms}
Simulation has become a more dominating approach with the increased accuracy and speed of simulation tools. National instruments' LabVIEW, which is a graphical computing platform, is widely used in the simulation program for testing and measurement \cite{arifin2022steering}. To model vehicle dynamics, the commercial tools Carsim \cite{rausch2017learning} with ADAMS \cite{sun2019nested}, CARLA \cite{del2021autonomous}, PreScan \cite{chae2017autonomous} and Matlab/Simulink \cite{chebly2019coupled} have been utilized by researchers for simulating car behavior. SCANeR Studio coupled with Matlab/Simulink is used to provide realistic driving conditions for testing of IV motion controllers \cite{chebly2019coupled}. Microscopic traffic simulation tools VISSIM 10.0 \cite{so2020automated} and AutonoVi-Sim \cite{best2018autonovi} are utilized for modeling traffic situations and implementing IV control. It should be noticed that the model errors should be considered through the verification and validation process and an inaccurate model may result in infeasible evaluation performance.

\subsubsection{Vehicles Platforms}
Some researchers set simulation studies as a first step for performance validation and conduct experimental tests for further validation. In the literature, the experimental tests can be conducted by a small and lightweight vehicle \cite{system_2}, and an on-road vehicle \cite{ arifin2022steering}. A small and lightweight vehicle is rugged and low-cost and can be carried by a single person for testing in various environmental conditions. In addition, the small and lightweight vehicle can travel at a high speed without the worry of injuring people and property. The main disadvantage is its small payload, which is insufficient to hold the computing power required for computing. On-road vehicles as Fig.  \ref{fig:system_fig_0} in contrast, can validate the control performance in diverse lighting and weather conditions, but the approval of testing such as ethical approval is needed.

\subsection{Testing Methods}
As the research progresses, researchers gradually build a framework for conducting research. They summarize effective methods and focused on several key works \cite{testing_tiv}.

\subsubsection{Testing Partition}
Researchers divide the testing into three sub-tasks, including function, performance, and intelligence testing. Function testing focuses on whether the vehicle or its component can obtain the required output for a specific input under given particular conditions (for example, can the Lidar give an output of more than 3 cloud points for a 1m*1m object located 200m away?). Performance testing focuses on whether the vehicle can obtain the required output for a specific range of inputs under a given range of conditions. Intelligence testing further requires vehicles to perform reasonable and intelligent perception, planning, and control strategies for broader and more abstract driving scenarios \cite{testing_add_1}.

\subsubsection{Testing Scenarios}
For the intelligence testing of IVs, currently, most of the researchers conduct analysis from the view of behavioral intelligence of artificial intelligence. They believe that if an IV drives successfully and smoothly under a given traffic scenario, then it is equipped with driving intelligence of such a scenario. As a consequence, it is a hot research topic on how to determine the necessary scenarios for testing. Addressing this need, the International Organization for Standardization proposes an international standard for functional safety (ISO 26262) aiming to point out what functional outputs should the vehicle obtain for a particular scenario with specific inputs. Furthermore, in order to reduce the dangers of unacceptable risks due to a lack of relative functions of the system (inadequate design or performance limitation) or foreseeable human error, the International Organization for Standardization proposes the estimated functional safety standard (ISO 21448) for functional testing of those functions that are affected by the external environment, such as automatic emergency braking (AEB) system, lane keeping system and other advanced driving assistance systems (ADAS). However, for the intelligence testing of IVs, the necessary scenarios are numerous. So far, it is hard to efficiently and reliably test the intelligence of vehicles by the method of finding the ideal relationship between inputs and outputs based on human expert-specified testing scenarios \cite{testing_add_2}.

\subsubsection{Learning-based Approaches}
Considering the difficulty of constructing scenarios and the trend of data-driven machine learning modeling in most artificial intelligence systems, researchers propose several algorithms to generate scenarios semi-automated or fully automated by learning from naturalistic driving data. Especially, researchers are strongly interested in two problems. One of the problems is how to quickly identify difficult testing scenarios, so as to avoid wasting time in simple testing scenarios which does not help improve vehicle performance \cite{testing_add_3, testing_add_4, testing_add_5}. The other problem is how to cover all possible scenarios as much as possible, so as to avoid IV encountering untested scenarios that are hard to deal with \cite{testing_add_6, testing_add_7}. Apparently, it is difficult to balance those two problems, and it remains unsolved. It is helpful to reveal the intelligence limitation of IVs by finding corner cases, which is worth to be studied further \cite{testing_add_8}.

\subsubsection{Parallel Testing}
As it is time-consuming to conduct real tests, most researchers focus on virtual tests based on simulation \cite{testing_add_9, testing_add_10, testing_add_11}. The experience of some famous companies such as Waymo \cite{testing_add_12} and Nvidia \cite{testing_add_13} shows that well-designed virtual tests can effectively identify the weakness of vehicles and provide useful modification solutions. Currently, relevant research is focused on how to ensure the intrinsic behavioral rationality and external manifestation variety of simulated objects in simulation systems. For example, some researchers discuss how to mimic human drivers’ behavior by learning from daily collected driving data, so that interactions between manned and unmanned vehicles can be correctly reproduced in simulation systems \cite{testing_add_14}. How to transform images by deep learning models, and obtain rarely seen special scene images from transforming daily driving collected scene images has been favored by many researchers \cite{testing_add_15, testing_add_16}.

So far there are still many difficulties that need to be solved in the field of IVs testing. It can be forecasted that the development of testing will drive into a crucial stage in the next decade.

%% file: body/methodology/human-mechine.tex
\section{Human Behaviors in Intelligent Vehicles}

Human behaviors and human factor issues are important topics for IVs, as they determine the widespread acceptance of IVs. Modeling and understanding human behaviors can also contribute to the establishment of mutual understanding and mutual trust between humans and vehicles. However, human is a highly complex system and a unified human behavior modeling framework still need to be investigated further \cite{human-machine_1}. In this section, the human behavior and human factor issues will be discussed according to the level of autonomy (SAE J3016) as the differences in human behavior can be significant \cite{human-machine_2}. 

\subsection{Driver Assistance and Partial Driving Automation (L1-L2)}
On L1 or L2 AD vehicles, human drivers are still inside the vehicle control loop and suppose to drive (mental drive) even if the feet or hands are off the pedals and steering wheel. In this case, the human driver can benefit from a series of ADAS such as lane departure warning (LDW) \cite{human-machine_3}, lane departure assistant (LDA) \cite{human-machine_4}, adaptive cruise control (ACC) \cite{human-machine_5}, and lane-keeping assistance (LKA) \cite{human-machine_6}, etc. To provide accurate assistance to human drivers, several popular research areas were developed, including fatigue (muscle/mental)/drowsiness detection \cite{human-machine_7}, driver intention inference \cite{human-machine_8}, attention \cite{human-machine_9}, workload \cite{human-machine_10}, emotion \cite{human-machine_11}, and distractions \cite{human-machine_12}. Such assistance systems usually can be separated into two fundamental groups, which are physical (physiological)-behavior and psychological behavior-based methods.


\subsection{Conditional Driving Automation (L3)}
On L3 AD vehicles, a human driver can further engage in secondary tasks, and long-term monitoring is no longer required. The human driver is only required to be present and capable of taking over the control at any time, especially if an emergency occurs due to system malfunction and surrounding context uncertainty. However, one of the famous drawbacks of L3 automation is the human driver cannot guarantee to provide quality and safe take-over control especially facing the time-constrained decision-making situation \cite{human-machine_16}. Therefore, the most widely studied points for L3 regarding human factors issues are the design of take-over control algorithms, resuming control, and smooth switching of the control authority \cite{human-machine_17}. 

\subsection{High/Fully Driving Automation (L4-L5)}
With higher-level automation, human drivers are no longer expected to maintain a proper situation awareness level in case to control the vehicle when necessary. However, complex issues can arise as well, one of which is how humans understand and partner with the autonomous system. To enable acceptance and behavior adaptation to higher-level IVs, driver trust should be developed. 

Driver trust is currently a major reason for the generalization and commercialization of AD and IVs. As shown in \cite{human-machine_23}, human trust in the IVs largely depends on the automation's driving performance considering the safety, comfortability, predictability, and ethical response to critical situations. A recent study \cite{human-machine_24} supported that the initial introduction of self-driving cars will cause unexpected reactions and situations, which may affect the level of trust the public has in the new technique. Choi et al. surveyed 552 drivers on their attitudes toward AD; it was shown that trust in autonomy is the most dominant factor \cite{human-machine_25}. According to an investigation of 162 Tesla drivers on their trust in Autopilot and Summon systems \cite{human-machine_26}, it was found that high levels of initial trust in the systems could be established if an excellent introduction to the system capability is provided, and the frequency of system usage will increase over time, regardless of whether participants experienced an automation failure. 

\subsection{Human-Machine-Interface (HMI)}
The AD techniques make the design of HMI even more complex as the HMI system for AD vehicles relies on human-centered design approaches \cite{human-machine_27} to ensure safety awareness, mutual trust, pleasure, and comfort \cite{human-machine_28}. Currently, the HMI system becomes the main module that enables the collaboration between driver and vehicle. The design of the HMI system is a system engineering task \cite{human-machine_29}, and this section focuses on the introduction of in-cabin and out-cabin HMI design. 

For L4/L5, vehicle autonomy is expected to be fully responsible for the driving tasks, and human drivers are even no longer required to maintain driving knowledge and skills. Hence, it is hard to require the human occupier to understand the context properly. Instead, only safety-related information for the vehicle behaviors and future planning has to be reported and displayed \cite{human-machine_30}. A satisfactory classification for various types of HMI on highly AD vehicles was proposed in \cite{human-machine_31}. Specifically, the onboard HMI system can be divided into five categories, namely, dynamic HMI, automation HMI, information HMI, vehicle HMI, and external HMI. Three recommended HMI design factors were introduced in \cite{human-machine_32}. First, a head-up display can avoid the distraction of the human driver. Second, HMI systems should assist with time-constrained maneuvers and decisions. Last, the driver monitoring system is needed to help understand the human states for high-quality visualization and explanation. An expert evaluation study for the design of HMI was discussed in \cite{human-machine_33}. According to the expert evaluation, the two most essential functions of HMI are 1) to present the automation availability, navigation, and environmental information, and 2) to improve the driver's attention through speech instruction, LED bar light, and seat oscillation. 

\subsection{Intelligent Cockpit Systems}
Intelligent cockpit (IC) design for IVs is another emerging topic in recent years. Indeed, some of the conventional IC techniques for low autonomy vehicles, such as driving behavior monitoring techniques can be adapted \cite{IC_1}. However, more challenging technologies are keen to be developed in case to satisfy the stricter requirement in safety, security, comfortability, HMI, and entertainment on IVs.  Regarding the safety issue, as discussed in \cite{human-machine_2}, an efficient mutual communication mechanism should be developed for IVs in case human intervention is needed in a critical situation. For full IVs, such information exchange is also important in the IC so that human passengers maintain proper situational awareness and trust in the autonomy by visualizing more time-critical traffic context information \cite{survey_3}. As more advanced vehicular technologies, such as V2X, blockchain, and federated learning have been introduced to IVs \cite{IC_4}, IVs are evolving into mobile information centers. In this case, cyber-security proposed another dramatic challenge to the IC as it is the essential requirement to protect the users’ privacy and safety to ensure a seamless journey and all the onboard technologies satisfy the ethical and policy issues. The comfortable consideration for IC on IVs is also different from that on conventional vehicles. It was shown that HMI has the greatest impact on comfortability, followed by the thermal environment, acoustic environment, and optical environment \cite{IC_5}. Similar conclusions are also developed from recent human motion sickness studies on IVs \cite{IC_6} as it is more likely to get motion sickness on IVs and HMI can mitigate such issues \cite{IC_7}. In sum, IC on IVs should not only consider the entertainment functions but need to address a series of safety, security, comfortability, human factor, and even energy consumption challenges that are caused by the nature of IVs.

%% file: body/conclusion.tex
\section{conclusion}
This article is the second part of our work(Part \uppercase\expandafter{\romannumeral1} for the technology survey). In this paper, we provide a review of wide introductions on research development with milestones in AD and IVs. In addition, we introduce plenty of deployment details, testing methods, and unique opinions. In combination with the other two parts, we expect that our whole work will bring novel and diverse insights to researchers and abecedarians, and serve as a bridge between past and future.

%% file: body/bib.tex
\small\small\bibliographystyle{IEEEtran}
\bibliography{body/mylib.bib}

%% file: main.bbl
\begin{thebibliography}{100}
\providecommand{\url}[1]{#1}
\csname url@samestyle\endcsname
\providecommand{\newblock}{\relax}
\providecommand{\bibinfo}[2]{#2}
\providecommand{\BIBentrySTDinterwordspacing}{\spaceskip=0pt\relax}
\providecommand{\BIBentryALTinterwordstretchfactor}{4}
\providecommand{\BIBentryALTinterwordspacing}{\spaceskip=\fontdimen2\font plus
\BIBentryALTinterwordstretchfactor\fontdimen3\font minus
  \fontdimen4\font\relax}
\providecommand{\BIBforeignlanguage}[2]{{%
\expandafter\ifx\csname l@#1\endcsname\relax
\typeout{** WARNING: IEEEtran.bst: No hyphenation pattern has been}%
\typeout{** loaded for the language `#1'. Using the pattern for}%
\typeout{** the default language instead.}%
\else
\language=\csname l@#1\endcsname
\fi
#2}}
\providecommand{\BIBdecl}{\relax}
\BIBdecl

\bibitem{SOS}
L.~Chen, Y.~Li, C.~Huang, B.~Li, Y.~Xing, D.~Tian, L.~Li, Z.~Hu, X.~Na, Z.~Li,
  S.~Teng, C.~Lv, J.~Wang, D.~Cao, N.~Zheng, and F.-Y. Wang, ``Milestones in
  autonomous driving and intelligent vehicles: Survey of surveys,'' \emph{IEEE
  Transactions on Intelligent Vehicles}, vol.~8, no.~2, pp. 1046--1056, 2023.

\bibitem{survey_control_6}
S.~Kuutti, R.~Bowden, Y.~Jin, P.~Barber, and S.~Fallah, ``A survey of deep
  learning applications to autonomous vehicle control,'' \emph{IEEE
  Transactions on Intelligent Transportation Systems}, vol.~22, no.~2, pp.
  712--733, 2020.

\bibitem{arefizadeh2018platooning}
S.~Arefizadeh and A.~Talebpour, ``A platooning strategy for automated vehicles
  in the presence of speed limit fluctuations,'' \emph{Transportation Research
  Record}, vol. 2672, no.~20, pp. 154--161, 2018.

\bibitem{malikopoulos2018optimal}
A.~A. Malikopoulos, S.~Hong, B.~B. Park, J.~Lee, and S.~Ryu, ``Optimal control
  for speed harmonization of automated vehicles,'' \emph{IEEE Transactions on
  Intelligent Transportation Systems}, vol.~20, no.~7, pp. 2405--2417, 2018.

\bibitem{li2018piecewise}
X.~Li, A.~Ghiasi, Z.~Xu, and X.~Qu, ``A piecewise trajectory optimization model
  for connected automated vehicles: Exact optimization algorithm and queue
  propagation analysis,'' \emph{Transportation Research Part B:
  Methodological}, vol. 118, pp. 429--456, 2018.

\bibitem{hao2018eco}
P.~Hao, G.~Wu, K.~Boriboonsomsin, and M.~J. Barth, ``Eco-approach and departure
  ({EAD}) application for actuated signals in real-world traffic,'' \emph{IEEE
  Transactions on Intelligent Transportation Systems}, vol.~20, no.~1, pp.
  30--40, 2018.

\bibitem{kim1996fuzzy}
H.~M. Kim, J.~Dickerson, and B.~Kosko, ``Fuzzy throttle and brake control for
  platoons of smart cars,'' \emph{Fuzzy Sets and Systems}, vol.~84, no.~3, pp.
  209--234, 1996.

\bibitem{choi1996robust}
S.-B. Choi and J.~Hedrick, ``Robust throttle control of automotive engines:
  Theory and experiment,'' 1996.

\bibitem{zhou2019distributed}
Y.~Zhou, M.~Wang, and S.~Ahn, ``Distributed model predictive control approach
  for cooperative car-following with guaranteed local and string stability,''
  \emph{Transportation Research Part B: Methodological}, vol. 128, pp. 69--86,
  2019.

\bibitem{huang2019game}
K.~Huang, X.~Di, Q.~Du, and X.~Chen, ``A game-theoretic framework for
  autonomous vehicles velocity control: Bridging microscopic differential games
  and macroscopic mean field games,'' \emph{arXiv preprint arXiv:1903.06053},
  2019.

\bibitem{wu2019path}
Y.~Wu, L.~Wang, J.~Zhang, and F.~Li, ``Path following control of autonomous
  ground vehicle based on nonsingular terminal sliding mode and active
  disturbance rejection control,'' \emph{IEEE Transactions on Vehicular
  Technology}, vol.~68, no.~7, pp. 6379--6390, 2019.

\bibitem{mar2001anfis}
J.~Mar and F.-J. Lin, ``An {ANFIS} controller for the car-following collision
  prevention system,'' \emph{IEEE Transactions on Vehicular Technology},
  vol.~50, no.~4, pp. 1106--1113, 2001.

\bibitem{swaroop2001direct}
D.~Swaroop, J.~K. Hedrick, and S.~B. Choi, ``Direct adaptive longitudinal
  control of vehicle platoons,'' \emph{IEEE Transactions on Vehicular
  Technology}, vol.~50, no.~1, pp. 150--161, 2001.

\bibitem{chen2017learning}
X.~Chen, Y.~Zhai, C.~Lu, J.~Gong, and G.~Wang, ``A learning model for
  personalized adaptive cruise control,'' in \emph{2017 IEEE Intelligent
  Vehicles Symposium (IV)}.\hskip 1em plus 0.5em minus 0.4em\relax IEEE, 2017,
  pp. 379--384.

\bibitem{dai2005approach}
X.~Dai, C.-K. Li, and A.~B. Rad, ``An approach to tune fuzzy controllers based
  on reinforcement learning for autonomous vehicle control,'' \emph{IEEE
  Transactions on Intelligent Transportation Systems}, vol.~6, no.~3, pp.
  285--293, 2005.

\bibitem{MotionPlanning}
S.~Teng, X.~Hu, P.~Deng, B.~Li, Y.~Li, Y.~Ai, D.~Yang, L.~Li, Z.~Xuanyuan,
  F.~Zhu, and L.~Chen, ``Motion planning for autonomous driving: The state of
  the art and future perspectives,'' \emph{IEEE Transactions on Intelligent
  Vehicles}, pp. 1--21, 2023.

\bibitem{gong2016constrained}
S.~Gong, J.~Shen, and L.~Du, ``Constrained optimization and distributed
  computation based car following control of a connected and autonomous vehicle
  platoon,'' \emph{Transportation Research Part B: Methodological}, vol.~94,
  pp. 314--334, 2016.

\bibitem{survey_planning_4}
L.~Claussmann, M.~Revilloud, D.~Gruyer, and S.~Glaser, ``A review of motion
  planning for highway autonomous driving,'' \emph{IEEE Transactions on
  Intelligent Transportation Systems}, vol.~21, no.~5, pp. 1826--1848, 2019.

\bibitem{zhang2019game}
Q.~Zhang, R.~Langari, H.~E. Tseng, D.~Filev, S.~Szwabowski, and S.~Coskun, ``A
  game theoretic model predictive controller with aggressiveness estimation for
  mandatory lane change,'' \emph{IEEE Transactions on Intelligent Vehicles},
  vol.~5, no.~1, pp. 75--89, 2019.

\bibitem{marino2011nested}
R.~Marino, S.~Scalzi, and M.~Netto, ``Nested {PID} steering control for lane
  keeping in autonomous vehicles,'' \emph{Control Engineering Practice},
  vol.~19, no.~12, pp. 1459--1467, 2011.

\bibitem{kim2014model}
E.~Kim, J.~Kim, and M.~Sunwoo, ``Model predictive control strategy for smooth
  path tracking of autonomous vehicles with steering actuator dynamics,''
  \emph{International Journal of Automotive Technology}, vol.~15, no.~7, pp.
  1155--1164, 2014.

\bibitem{ganzelmeier2001robustness}
L.~Ganzelmeier, J.~Helbig, and E.~Schnieder, ``Robustness and performance
  advanced control of vehicle dynamics,'' in \emph{IEEE Int'l Transp Conf
  Proc}, 2001, pp. 25--29.

\bibitem{netto2004lateral}
M.~S. Netto, S.~Chaib, and S.~Mammar, ``Lateral adaptive control for vehicle
  lane keeping,'' in \emph{Proceedings of the 2004 American Control
  Conference}, vol.~3.\hskip 1em plus 0.5em minus 0.4em\relax IEEE, 2004, pp.
  2693--2698.

\bibitem{besselmann2009autonomous}
T.~Besselmann and M.~Morari, ``Autonomous vehicle steering using explicit
  {LPV-MPC},'' in \emph{2009 European Control Conference (ECC)}.\hskip 1em plus
  0.5em minus 0.4em\relax IEEE, 2009, pp. 2628--2633.

\bibitem{liu2021super}
J.~Liu, L.~Gao, J.~Zhang, and F.~Yan, ``Super-twisting algorithm second-order
  sliding mode control for collision avoidance system based on active front
  steering and direct yaw moment control,'' \emph{Proceedings of the
  Institution of Mechanical Engineers, Part D: Journal of Automobile
  Engineering}, vol. 235, no.~1, pp. 43--54, 2021.

\bibitem{arifin2022steering}
B.~Arifin, B.~Y. Suprapto, S.~A.~D. Prasetyowati, and Z.~Nawawi, ``Steering
  control in electric power steering autonomous vehicle using type-2 fuzzy
  logic control and pi control,'' \emph{World Electric Vehicle Journal},
  vol.~13, no.~3, p.~53, 2022.

\bibitem{eraqi2017end}
H.~M. Eraqi, M.~N. Moustafa, and J.~Honer, ``End-to-end deep learning for
  steering autonomous vehicles considering temporal dependencies,'' \emph{arXiv
  preprint arXiv:1710.03804}, 2017.

\bibitem{yu2018human}
H.~Yu, H.~E. Tseng, and R.~Langari, ``A human-like game theory-based controller
  for automatic lane changing,'' \emph{Transportation Research Part C: Emerging
  Technologies}, vol.~88, pp. 140--158, 2018.

\bibitem{huang2017parameterized}
Z.~Huang, X.~Xu, H.~He, J.~Tan, and Z.~Sun, ``Parameterized batch reinforcement
  learning for longitudinal control of autonomous land vehicles,'' \emph{IEEE
  Transactions on Systems, Man, and Cybernetics: Systems}, vol.~49, no.~4, pp.
  730--741, 2017.

\bibitem{cai2009intelligent}
L.~Cai, A.~B. Rad, and W.-L. Chan, ``An intelligent longitudinal controller for
  application in semiautonomous vehicles,'' \emph{IEEE Transactions on
  Industrial Electronics}, vol.~57, no.~4, pp. 1487--1497, 2009.

\bibitem{sun2019nested}
Z.~Sun, J.~Zheng, Z.~Man, M.~Fu, and R.~Lu, ``Nested adaptive super-twisting
  sliding mode control design for a vehicle steer-by-wire system,''
  \emph{Mechanical Systems and Signal Processing}, vol. 122, pp. 658--672,
  2019.

\bibitem{guo2012design}
J.~Guo, P.~Hu, L.~Li, and R.~Wang, ``Design of automatic steering controller
  for trajectory tracking of unmanned vehicles using genetic algorithms,''
  \emph{IEEE Transactions on Vehicular Technology}, vol.~61, no.~7, pp.
  2913--2924, 2012.

\bibitem{8885839}
O.~Pauca, C.~F. Caruntu, and C.~Lazar, ``Predictive control for the lateral and
  longitudinal dynamics in automated vehicles,'' in \emph{2019 23rd
  International Conference on System Theory, Control and Computing (ICSTCC)},
  2019, pp. 797--802.

\bibitem{nobe2001overview}
S.~A. Nobe and F.-Y. Wang, ``An overview of recent developments in automated
  lateral and longitudinal vehicle controls,'' in \emph{2001 IEEE International
  Conference on Systems, Man and Cybernetics. E-Systems and E-Man for
  Cybernetics in Cyberspace (Cat. No. 01CH37236)}, vol.~5.\hskip 1em plus 0.5em
  minus 0.4em\relax IEEE, 2001, pp. 3447--3452.

\bibitem{rausch2017learning}
V.~Rausch, A.~Hansen, E.~Solowjow, C.~Liu, E.~Kreuzer, and J.~K. Hedrick,
  ``Learning a deep neural net policy for end-to-end control of autonomous
  vehicles,'' in \emph{2017 American Control Conference (ACC)}.\hskip 1em plus
  0.5em minus 0.4em\relax IEEE, 2017, pp. 4914--4919.

\bibitem{gaspar2005design}
P.~G{\'a}sp{\'a}r, Z.~Szab{\'o}, and J.~Bokor, ``The design of an integrated
  control system in heavy vehicles based on an {LPV} method,'' in
  \emph{Proceedings of the 44th IEEE Conference on Decision and Control}.\hskip
  1em plus 0.5em minus 0.4em\relax IEEE, 2005, pp. 6722--6727.

\bibitem{Hierarchical}
S.~Teng, L.~Chen, Y.~Ai, Y.~Zhou, Z.~Xuanyuan, and X.~Hu, ``Hierarchical
  interpretable imitation learning for end-to-end autonomous driving,''
  \emph{IEEE Transactions on Intelligent Vehicles}, vol.~8, no.~1, pp.
  673--683, 2023.

\bibitem{yu1995road}
G.~Yu and I.~K. Sethi, ``Road-following with continuous learning,'' in
  \emph{Proceedings of the Intelligent Vehicles' 95. Symposium}.\hskip 1em plus
  0.5em minus 0.4em\relax IEEE, 1995, pp. 412--417.

\bibitem{wang2018reinforcement}
P.~Wang, C.-Y. Chan, and A.~de~La~Fortelle, ``A reinforcement learning based
  approach for automated lane change maneuvers,'' in \emph{2018 IEEE
  Intelligent Vehicles Symposium (IV)}.\hskip 1em plus 0.5em minus 0.4em\relax
  IEEE, 2018, pp. 1379--1384.

\bibitem{chae2017autonomous}
H.~Chae, C.~M. Kang, B.~Kim, J.~Kim, C.~C. Chung, and J.~W. Choi, ``Autonomous
  braking system via deep reinforcement learning,'' in \emph{2017 IEEE 20th
  International Conference on Intelligent Transportation Systems (ITSC)}.\hskip
  1em plus 0.5em minus 0.4em\relax IEEE, 2017, pp. 1--6.

\bibitem{naranjo2006acc+}
J.~E. Naranjo, C.~Gonz{\'a}lez, R.~Garc{\'\i}a, and T.~De~Pedro, ``{ACC}+
  stop\&go maneuvers with throttle and brake fuzzy control,'' \emph{IEEE
  Transactions on Intelligent Transportation Systems}, vol.~7, no.~2, pp.
  213--225, 2006.

\bibitem{samak2011control}
C.~V. Samak, T.~V. Samak, and S.~Kandhasamy, ``Control strategies for
  autonomous vehicles,'' in \emph{Autonomous Driving and Advanced
  Driver-Assistance Systems (ADAS)}.\hskip 1em plus 0.5em minus 0.4em\relax CRC
  Press, 2021, pp. 37--86.

\bibitem{survey_8}
L.~Liu, S.~Lu, R.~Zhong, B.~Wu, Y.~Yao, Q.~Zhang, and W.~Shi, ``Computing
  systems for autonomous driving: State of the art and challenges,'' \emph{IEEE
  Internet of Things Journal}, vol.~8, no.~8, pp. 6469--6486, 2020.

\bibitem{system_2}
U.~Muller, J.~Ben, E.~Cosatto, B.~Flepp, and Y.~Cun, ``Off-road obstacle
  avoidance through end-to-end learning,'' \emph{Advances in Neural Information
  Processing Systems}, vol.~18, 2005.

\bibitem{survey_system_1}
S.-C. Lin, Y.~Zhang, C.-H. Hsu, M.~Skach, M.~E. Haque, L.~Tang, and J.~Mars,
  ``The architectural implications of autonomous driving: Constraints and
  acceleration,'' in \emph{Proceedings of the Twenty-Third International
  Conference on Architectural Support for Programming Languages and Operating
  Systems}, 2018, pp. 751--766.

\bibitem{survey_system_2}
S.~Liu, L.~Liu, J.~Tang, B.~Yu, Y.~Wang, and W.~Shi, ``Edge computing for
  autonomous driving: Opportunities and challenges,'' \emph{Proceedings of the
  IEEE}, vol. 107, no.~8, pp. 1697--1716, 2019.

\bibitem{system_9}
P.~K.~D. Jagannadha, M.~Yilmaz, M.~Sonawane, S.~Chadalavada, S.~Sarangi,
  B.~Bhaskaran, S.~Bajpai, V.~A. Reddy, J.~Pandey, and S.~Jiang, ``Special
  session: in-system-test ({IST}) architecture for {N}vidia drive-{AGX}
  platforms,'' in \emph{2019 IEEE 37th VLSI Test Symposium (VTS)}.\hskip 1em
  plus 0.5em minus 0.4em\relax IEEE, 2019, pp. 1--8.

\bibitem{system_10}
V.~Boppana, S.~Ahmad, I.~Ganusov, V.~Kathail, V.~Rajagopalan, and R.~Wittig,
  ``Ultrascale+ {MPS}o{C} and {FPGA} families,'' in \emph{2015 IEEE Hot Chips
  27 Symposium (HCS)}.\hskip 1em plus 0.5em minus 0.4em\relax IEEE, 2015, pp.
  1--37.

\bibitem{system_11}
N.~P. Jouppi, D.~H. Yoon, M.~Ashcraft, M.~Gottscho, T.~B. Jablin, G.~Kurian,
  J.~Laudon, S.~Li, P.~Ma, X.~Ma \emph{et~al.}, ``Ten lessons from three
  generations shaped {G}oogle’s {TPU}v4i: Industrial product,'' in \emph{2021
  ACM/IEEE 48th Annual International Symposium on Computer Architecture
  (ISCA)}.\hskip 1em plus 0.5em minus 0.4em\relax IEEE, 2021, pp. 1--14.

\bibitem{OS}
L.~Chen, Y.~Zhang, B.~Tian, Y.~Ai, D.~Cao, and F.-Y. Wang, ``Parallel driving
  os: A ubiquitous operating system for autonomous driving in cpss,''
  \emph{IEEE Transactions on Intelligent Vehicles}, vol.~7, no.~4, pp.
  886--895, 2022.

\bibitem{system_12}
D.~Hildebrand, ``An architectural overview of {QNX}.'' in \emph{USENIX Workshop
  on Microkernels and Other Kernel Architectures}, 1992, pp. 113--126.

\bibitem{system_13}
P.~Hambarde, R.~Varma, and S.~Jha, ``The survey of real time operating system:
  {RTOS},'' in \emph{2014 International Conference on Electronic Systems,
  Signal Processing and Computing Technologies}.\hskip 1em plus 0.5em minus
  0.4em\relax IEEE, 2014, pp. 34--39.

\bibitem{system_7}
T.~Wu, B.~Wu, S.~Wang, L.~Liu, S.~Liu, Y.~Bao, and W.~Shi, ``Oops! it's too
  late. your autonomous driving system needs a faster middleware,'' \emph{IEEE
  Robotics and Automation Letters}, vol.~6, no.~4, pp. 7301--7308, 2021.

\bibitem{system_15}
M.~Quigley, K.~Conley, B.~Gerkey, J.~Faust, T.~Foote, J.~Leibs, R.~Wheeler,
  A.~Y. Ng \emph{et~al.}, ``{ROS}: an open-source robot operating system,'' in
  \emph{ICRA Workshop on Open Source Software}, vol.~3, no. 3.2.\hskip 1em plus
  0.5em minus 0.4em\relax Kobe, Japan, 2009, p.~5.

\bibitem{system_16}
V.~DiLuoffo, W.~R. Michalson, and B.~Sunar, ``Robot operating system 2: The
  need for a holistic security approach to robotic architectures,''
  \emph{International Journal of Advanced Robotic Systems}, vol.~15, no.~3, p.
  1729881418770011, 2018.

\bibitem{system_17}
M.~A. Fayazbakhsh and M.~Bahrami, ``Comprehensive modeling of vehicle air
  conditioning loads using heat balance method,'' \emph{SAE Technical Paper},
  vol. 2013, p. 1507, 2013.

\bibitem{survey_hardware_3}
D.~J. Yeong, G.~Velasco-Hernandez, J.~Barry, and J.~Walsh, ``Sensor and sensor
  fusion technology in autonomous vehicles: A review,'' \emph{Sensors},
  vol.~21, no.~6, p. 2140, 2021.

\bibitem{survey_dynamic_15}
J.~Fayyad, M.~A. Jaradat, D.~Gruyer, and H.~Najjaran, ``Deep learning sensor
  fusion for autonomous vehicle perception and localization: A review,''
  \emph{Sensors}, vol.~20, no.~15, p. 4220, 2020.

\bibitem{survey_dynamic_16}
Y.~Li, L.~Ma, Z.~Zhong, F.~Liu, M.~A. Chapman, D.~Cao, and J.~Li, ``Deep
  learning for lidar point clouds in autonomous driving: A review,'' \emph{IEEE
  Transactions on Neural Networks and Learning Systems}, vol.~32, no.~8, pp.
  3412--3432, 2020.

\bibitem{system_27}
S.~Royo and M.~Ballesta-Garcia, ``An overview of lidar imaging systems for
  autonomous vehicles,'' \emph{Applied Sciences}, vol.~9, no.~19, p. 4093,
  2019.

\bibitem{system_24}
D.~Wang, C.~Watkins, and H.~Xie, ``{MEMS} mirrors for {L}i{DAR}: a review,''
  \emph{Micromachines}, vol.~11, no.~5, p. 456, 2020.

\bibitem{survey_dynamic_21}
Y.~Cui, R.~Chen, W.~Chu, L.~Chen, D.~Tian, Y.~Li, and D.~Cao, ``Deep learning
  for image and point cloud fusion in autonomous driving: A review,''
  \emph{IEEE Transactions on Intelligent Transportation Systems}, vol.~23,
  no.~2, pp. 722--739, 2021.

\bibitem{survey_hardware_2}
T.~Zhou, M.~Yang, K.~Jiang, H.~Wong, and D.~Yang, ``Mmw radar-based
  technologies in autonomous driving: A review,'' \emph{Sensors}, vol.~20,
  no.~24, p. 7283, 2020.

\bibitem{system_35}
D.~Ma, N.~Shlezinger, T.~Huang, Y.~Liu, and Y.~C. Eldar, ``Joint
  radar-communication strategies for autonomous vehicles: Combining two key
  automotive technologies,'' \emph{IEEE Signal Processing Magazine}, vol.~37,
  no.~4, pp. 85--97, 2020.

\bibitem{system_36}
S.~Saponara, M.~S. Greco, and F.~Gini, ``Radar-on-chip/in-package in autonomous
  driving vehicles and intelligent transport systems: Opportunities and
  challenges,'' \emph{IEEE Signal Processing Magazine}, vol.~36, no.~5, pp.
  71--84, 2019.

\bibitem{system_37}
S.~H. Javadi and A.~Farina, ``Radar networks: A review of features and
  challenges,'' \emph{Information Fusion}, vol.~61, pp. 48--55, 2020.

\bibitem{system_38}
H.~Mousazadeh, ``A technical review on navigation systems of agricultural
  autonomous off-road vehicles,'' \emph{Journal of Terramechanics}, vol.~50,
  no.~3, pp. 211--232, 2013.

\bibitem{system_39}
H.~Jing, Y.~Gao, S.~Shahbeigi, and M.~Dianati, ``Integrity monitoring of
  {GNSS/INS} based positioning systems for autonomous vehicles:
  {S}tate-of-the-{A}rt and open challenges,'' \emph{IEEE Transactions on
  Intelligent Transportation Systems}, 2022.

\bibitem{system_40}
F.~Rosique, P.~J. Navarro, C.~Fern{\'a}ndez, and A.~Padilla, ``A systematic
  review of perception system and simulators for autonomous vehicles
  research,'' \emph{Sensors}, vol.~19, no.~3, p. 648, 2019.

\bibitem{system_41}
A.~S. Mohammed, A.~Amamou, F.~K. Ayevide, S.~Kelouwani, K.~Agbossou, and
  N.~Zioui, ``The perception system of intelligent ground vehicles in all
  weather conditions: A systematic literature review,'' \emph{Sensors},
  vol.~20, no.~22, p. 6532, 2020.

\bibitem{system_47}
``Bsi cav standards roadmap 2022, howpublished =
  {\url{https://www.bsigroup.com/globalassets/localfiles/en-gb/cav/cav-standards-roadmap/cav-standards-roadmap-2022.pdf}}.''

\bibitem{system_48}
Y.~Gao, Y.~Ai, B.~Tian, L.~Chen, J.~Wang, D.~Cao, and F.-Y. Wang, ``Parallel
  end-to-end autonomous mining: An iot-oriented approach,'' \emph{IEEE Internet
  of Things Journal}, vol.~7, no.~2, pp. 1011--1023, 2020.

\bibitem{system_51}
B.~Van~Meldert and L.~De~Boeck, ``Introducing autonomous vehicles in logistics:
  a review from a broad perspective,'' \emph{FEB Research Report KBI\_1618},
  2016.

\bibitem{survey_11}
D.~Omeiza, H.~Webb, M.~Jirotka, and L.~Kunze, ``Explanations in autonomous
  driving: A survey,'' \emph{IEEE Transactions on Intelligent Transportation
  Systems}, 2021.

\bibitem{survey_explainable}
{\'E}.~Zablocki, H.~Ben-Younes, P.~P{\'e}rez, and M.~Cord, ``Explainability of
  vision-based autonomous driving systems: Review and challenges,'' \emph{arXiv
  preprint arXiv:2101.05307}, 2021.

\bibitem{system_54}
S.~Atakishiyev, M.~Salameh, H.~Yao, and R.~Goebel, ``Explainable artificial
  intelligence for autonomous driving: A comprehensive overview and field guide
  for future research directions,'' \emph{arXiv preprint arXiv:2112.11561},
  2021.

\bibitem{hartenstein2008tutorial}
H.~Hartenstein and L.~Laberteaux, ``A tutorial survey on vehicular ad hoc
  networks,'' \emph{IEEE Communications Magazine}, vol.~46, no.~6, pp.
  164--171, 2008.

\bibitem{abboud2016interworking}
K.~Abboud, H.~A. Omar, and W.~Zhuang, ``Interworking of {DSRC} and cellular
  network technologies for {V2X} communications: A survey,'' \emph{IEEE
  Transactions on Vehicular Technology}, vol.~65, no.~12, pp. 9457--9470, 2016.

\bibitem{chen2020vision}
S.~Chen, J.~Hu, Y.~Shi, L.~Zhao, and W.~Li, ``A vision of {C-V2X}:
  Technologies, field testing, and challenges with chinese development,''
  \emph{IEEE Internet of Things Journal}, vol.~7, no.~5, pp. 3872--3881, 2020.

\bibitem{uzcategui2009wave}
R.~A. Uzc{\'a}tegui, A.~J. De~Sucre, and G.~Acosta-Marum, ``Wave: A tutorial,''
  \emph{IEEE Communications Magazine}, vol.~47, no.~5, pp. 126--133, 2009.

\bibitem{survey_communication_7}
S.~Zeadally, J.~Guerrero, and J.~Contreras, ``A tutorial survey on
  vehicle-to-vehicle communications,'' \emph{Telecommunication Systems},
  vol.~73, no.~3, pp. 469--489, 2020.

\bibitem{teixeira2014vehicular}
F.~A. Teixeira, V.~F. e~Silva, J.~L. Leoni, D.~F. Macedo, and J.~M. Nogueira,
  ``Vehicular networks using the {IEEE} 802.11 p standard: An experimental
  analysis,'' \emph{Vehicular Communications}, vol.~1, no.~2, pp. 91--96, 2014.

\bibitem{dibaei2022investigating}
M.~Dibaei, X.~Zheng, Y.~Xia, X.~Xu, A.~Jolfaei, A.~K. Bashir, U.~Tariq, D.~Yu,
  and A.~V. Vasilakos, ``Investigating the prospect of leveraging blockchain
  and machine learning to secure vehicular networks: A survey,'' \emph{IEEE
  Transactions on Intelligent Transportation Systems}, vol.~23, no.~2, pp.
  683--700, 2022.

\bibitem{fcc2020use}
\BIBentryALTinterwordspacing
``Use of the 5.850-5.925 {GH}z {B}and,'' FCC 20-164, Federal Communications
  Commission, 2020. [Online]. Available:
  \url{https://docs.fcc.gov/public/attachments/FCC-20-164A1.pdf}
\BIBentrySTDinterwordspacing

\bibitem{gyawali2021challenges}
S.~Gyawali, S.~Xu, Y.~Qian, and R.~Q. Hu, ``Challenges and solutions for
  cellular based {V2X} communications,'' \emph{IEEE Communications Surveys \&
  Tutorials}, vol.~23, no.~1, pp. 222--255, 2021.

\bibitem{hasan2020securing}
M.~Hasan, S.~Mohan, T.~Shimizu, and H.~Lu, ``Securing vehicle-to-everything
  ({V2X}) communication platforms,'' \emph{IEEE Transactions on Intelligent
  Vehicles}, vol.~5, no.~4, pp. 693--713, 2020.

\bibitem{chen2016lte}
S.~Chen, J.~Hu, Y.~Shi, and L.~Zhao, ``{LTE-V}: A {TD-LTE}-based {V2X} solution
  for future vehicular network,'' \emph{IEEE Internet of Things Journal},
  vol.~3, no.~6, pp. 997--1005, 2016.

\bibitem{naik2019ieee}
G.~Naik, B.~Choudhury, and J.-M. Park, ``{IEEE} 802.11 bd \& 5{G} {NR} {V2X}:
  Evolution of radio access technologies for {V2X} communications,'' \emph{IEEE
  Access}, vol.~7, pp. 70\,169--70\,184, 2019.

\bibitem{ahmed2018cooperative}
E.~Ahmed and H.~Gharavi, ``Cooperative vehicular networking: A survey,''
  \emph{IEEE Transactions on Intelligent Transportation Systems}, vol.~19,
  no.~3, pp. 996--1014, 2018.

\bibitem{liang2017vehicular}
L.~Liang, H.~Peng, G.~Y. Li, and X.~Shen, ``Vehicular communications: A
  physical layer perspective,'' \emph{IEEE Transactions on Vehicular
  Technology}, vol.~66, no.~12, pp. 10\,647--10\,659, 2017.

\bibitem{wang2019networking}
J.~Wang, J.~Liu, and N.~Kato, ``Networking and communications in autonomous
  driving: A survey,'' \emph{IEEE Communications Surveys \& Tutorials},
  vol.~21, no.~2, pp. 1243--1274, 2019.

\bibitem{arslan2017effects}
S.~Arslan and M.~Saritas, ``The effects of {OFDM} design parameters on the
  {V2X} communication performance: A survey,'' \emph{Vehicular Communications},
  vol.~7, pp. 1--6, 2017.

\bibitem{ghafoor2020millimeter}
K.~Z. Ghafoor, L.~Kong, S.~Zeadally, A.~S. Sadiq, G.~Epiphaniou, M.~Hammoudeh,
  A.~K. Bashir, and S.~Mumtaz, ``Millimeter-wave communication for internet of
  vehicles: Status, challenges, and perspectives,'' \emph{IEEE Internet of
  Things Journal}, vol.~7, no.~9, pp. 8525--8546, 2020.

\bibitem{memedi2021vehicular}
A.~Memedi and F.~Dressler, ``Vehicular visible light communications: A
  survey,'' \emph{IEEE Communications Surveys \& Tutorials}, vol.~23, no.~1,
  pp. 161--181, 2021.

\bibitem{paranjothi2020survey}
A.~Paranjothi, M.~S. Khan, and S.~Zeadally, ``A survey on congestion detection
  and control in connected vehicles,'' \emph{Ad Hoc Networks}, vol. 108, no.
  102277, pp. 1--17, 2020.

\bibitem{rashmi2021survey}
K.~Rashmi and R.~Patil, ``Survey on cross layer approach for robust
  communication in {VANET},'' \emph{Wireless Personal Communications}, vol.
  119, no.~4, pp. 3413--3434, 2021.

\bibitem{noor2022survey}
M.~Noor-A-Rahim, Z.~Liu, H.~Lee, G.~M.~N. Ali, D.~Pesch, and P.~Xiao, ``A
  survey on resource allocation in vehicular networks,'' \emph{IEEE
  Transactions on Intelligent Transportation Systems}, vol.~23, no.~2, pp.
  701--721, 2022.

\bibitem{le2021comprehensive}
T.~T.~T. Le and S.~Moh, ``Comprehensive survey of radio resource allocation
  schemes for 5{G} {V2X} communications,'' \emph{IEEE Access}, vol.~9, pp.
  123\,117--123\,133, 2021.

\bibitem{nguyen2021drl}
H.~T. Nguyen, M.~T. Nguyen, H.~T. Do, H.~T. Hua, and C.~V. Nguyen, ``Drl-based
  intelligent resource allocation for diverse {Q}o{S} in 5{G} and toward 6{G}
  vehicular networks: a comprehensive survey,'' \emph{Wireless Communications
  and Mobile Computing}, vol. 2021, no. 5051328, pp. 1--21, 2021.

\bibitem{jeong2021comprehensive}
J.~Jeong, Y.~Shen, T.~Oh, S.~C{\'e}spedes, N.~Benamar, M.~Wetterwald, and
  J.~H{\"a}rri, ``A comprehensive survey on vehicular networks for smart roads:
  A focus on {IP}-based approaches,'' \emph{Vehicular Communications}, vol.~29,
  no. 100334, pp. 1--29, 2021.

\bibitem{xia2021comprehensive}
Z.~Xia, J.~Wu, L.~Wu, Y.~Chen, J.~Yang, and P.~S. Yu, ``A comprehensive survey
  of the key technologies and challenges surrounding vehicular ad hoc
  networks,'' \emph{ACM Transactions on Intelligent Systems and Technology},
  vol.~12, no.~4, pp. 1--30, 2021.

\bibitem{sun2021applications}
Z.~Sun, Y.~Liu, J.~Wang, G.~Li, C.~Anil, K.~Li, X.~Guo, G.~Sun, D.~Tian, and
  D.~Cao, ``Applications of game theory in vehicular networks: A survey,''
  \emph{IEEE Communications Surveys \& Tutorials}, vol.~23, no.~4, pp.
  2660--2710, 2021.

\bibitem{belamri2021survey}
F.~Belamri, S.~Boulfekhar, and D.~Aissani, ``A survey on {Q}o{S} routing
  protocols in {V}ehicular {A}d {H}oc {N}etwork ({VANET}),''
  \emph{Telecommunication Systems}, vol.~78, no.~1, pp. 117--153, 2021.

\bibitem{kayarga2021study}
T.~Kayarga and S.~A. Kumar, ``A study on various technologies to solve the
  routing problem in {I}nternet of {V}ehicles ({I}o{V}),'' \emph{Wireless
  Personal Communications}, vol. 119, no.~1, pp. 459--487, 2021.

\bibitem{talpur2022machine}
A.~Talpur and M.~Gurusamy, ``Machine learning for security in vehicular
  networks: A comprehensive survey,'' \emph{IEEE Communications Surveys \&
  Tutorials}, vol.~24, no.~1, pp. 346--379, 2022.

\bibitem{HDmap_81}
K.~Wong and C.~Ellul, ``Using geometry-based metrics as part of
  fitness-for-purpose evaluations of 3{D} city models,'' vol.~4, pp. 129--136,
  2016.

\bibitem{HDmap_79}
M.~Javanmardi, E.~Javanmardi, Y.~Gu, and S.~Kamijo, ``Towards high-definition
  3{D} urban mapping: Road feature-based registration of mobile mapping systems
  and aerial imagery,'' \emph{Remote Sensing}, vol.~9, no.~10, p. 975, 2017.

\bibitem{HDmap_82}
J.~Murphy, Y.~Pao, and A.~Yuen, ``Map matching when the map is wrong: Efficient
  vehicle tracking on-and off-road for map learning,'' \emph{arXiv preprint
  arXiv:1809.09755}, 2018.

\bibitem{HDmap_80}
Y.~Yang and G.~Huang, ``Map-based localization under adversarial attacks,'' pp.
  775--790, 2020.

\bibitem{del2021autonomous}
J.~del Egido~Sierra, A.~Diaz, L.~M. Bergasa, R.~Barea, and M.~L{\'o}pez,
  ``Autonomous vehicle control in carla challenge,'' \emph{Transportation
  Research Procedia}, vol.~58, pp. 69--74, 2021.

\bibitem{chebly2019coupled}
A.~Chebly, R.~Talj, and A.~Charara, ``Coupled longitudinal/lateral controllers
  for autonomous vehicles navigation, with experimental validation,''
  \emph{Control Engineering Practice}, vol.~88, pp. 79--96, 2019.

\bibitem{so2020automated}
J.~J. So, J.~Kang, S.~Park, I.~Park, and J.~Lee, ``Automated emergency vehicle
  control strategy based on automated driving controls,'' \emph{Journal of
  Advanced Transportation}, vol. 2020, 2020.

\bibitem{best2018autonovi}
A.~Best, S.~Narang, L.~Pasqualin, D.~Barber, and D.~Manocha, ``Autonovi-sim:
  Autonomous vehicle simulation platform with weather, sensing, and traffic
  control,'' in \emph{Proceedings of the IEEE Conference on Computer Vision and
  Pattern Recognition Workshops}, 2018, pp. 1048--1056.

\bibitem{testing_tiv}
J.~Wang, X.~Wang, T.~Shen, Y.~Wang, L.~Li, Y.~Tian, H.~Yu, L.~Chen, J.~Xin,
  X.~Wu, N.~Zheng, and F.-Y. Wang, ``Parallel vision for long-tail
  regularization: Initial results from ivfc autonomous driving testing,''
  \emph{IEEE Transactions on Intelligent Vehicles}, vol.~7, no.~2, pp.
  286--299, 2022.

\bibitem{testing_add_1}
L.~Li, W.-L. Huang, Y.~Liu, N.-N. Zheng, and F.-Y. Wang, ``Intelligence testing
  for autonomous vehicles: A new approach,'' \emph{IEEE Transactions on
  Intelligent Vehicles}, vol.~1, no.~2, pp. 158--166, 2016.

\bibitem{testing_add_2}
J.~Sun, H.~Zhang, H.~Zhou, R.~Yu, and Y.~Tian, ``Scenario-based test automation
  for highly automated vehicles: A review and paving the way for systematic
  safety assurance,'' \emph{IEEE Transactions on Intelligent Transportation
  Systems}, 2021.

\bibitem{testing_add_3}
D.~Zhao, H.~Lam, H.~Peng, S.~Bao, D.~J. LeBlanc, K.~Nobukawa, and C.~S. Pan,
  ``Accelerated evaluation of automated vehicles safety in lane-change
  scenarios based on importance sampling techniques,'' \emph{IEEE transactions
  on intelligent transportation systems}, vol.~18, no.~3, pp. 595--607, 2016.

\bibitem{testing_add_4}
S.~Feng, Y.~Feng, C.~Yu, Y.~Zhang, and H.~X. Liu, ``Testing scenario library
  generation for connected and automated vehicles, part i: Methodology,''
  \emph{IEEE Transactions on Intelligent Transportation Systems}, vol.~22,
  no.~3, pp. 1573--1582, 2020.

\bibitem{testing_add_5}
S.~Feng, Y.~Feng, H.~Sun, S.~Bao, Y.~Zhang, and H.~X. Liu, ``Testing scenario
  library generation for connected and automated vehicles, part ii: Case
  studies,'' \emph{IEEE Transactions on Intelligent Transportation Systems},
  vol.~22, no.~9, pp. 5635--5647, 2020.

\bibitem{testing_add_6}
L.~Li, N.~Zheng, and F.-Y. Wang, ``A theoretical foundation of intelligence
  testing and its application for intelligent vehicles,'' \emph{IEEE
  Transactions on Intelligent Transportation Systems}, vol.~22, no.~10, pp.
  6297--6306, 2020.

\bibitem{testing_add_7}
J.~Ge, H.~Xu, J.~Zhang, Y.~Zhang, D.~Yao, and L.~Li, ``Heterogeneous driver
  modeling and corner scenarios sampling for automated vehicles testing,''
  \emph{Journal of advanced transportation}, vol. 2022, 2022.

\bibitem{testing_add_8}
J.~Breitenstein, J.-A. Term{\"o}hlen, D.~Lipinski, and T.~Fingscheidt, ``Corner
  cases for visual perception in automated driving: Some guidance on detection
  approaches,'' \emph{arXiv preprint arXiv:2102.05897}, 2021.

\bibitem{testing_add_9}
F.-Y. Wang, X.~Wang, L.~Li, and P.~Mirchandani, ``Creating a digital-vehicle
  proving ground,'' \emph{IEEE Intelligent systems}, vol.~18, no.~2, pp.
  12--15, 2003.

\bibitem{testing_add_10}
F.-Y. Wang, N.~Zheng, L.~Li, J.~Xin, X.~Wang, L.~Xu, B.~Tian, G.~Wu, Z.~Zhang,
  C.~Wang \emph{et~al.}, ``China's 12-year quest of autonomous vehicular
  intelligence: The intelligent vehicles future challenge program,'' \emph{IEEE
  Intelligent Transportation Systems Magazine}, vol.~13, no.~2, pp. 6--19,
  2021.

\bibitem{testing_add_11}
F.-Y. Wang, R.~Song, R.~Zhou, X.~Wang, L.~Chen, L.~Li, L.~Zeng, J.~Zhou,
  S.~Teng, and X.~Zhu, ``Verification and validation of intelligent vehicles:
  Objectives and efforts from china,'' \emph{IEEE Transactions on Intelligent
  Vehicles}, vol.~7, no.~2, pp. 164--169, 2022.

\bibitem{testing_add_12}
J.~M. Scanlon, K.~D. Kusano, T.~Daniel, C.~Alderson, A.~Ogle, and T.~Victor,
  ``Waymo simulated driving behavior in reconstructed fatal crashes within an
  autonomous vehicle operating domain,'' \emph{Accident Analysis \&
  Prevention}, vol. 163, p. 106454, 2021.

\bibitem{testing_add_13}
D.~Acuna, J.~Philion, and S.~Fidler, ``Towards optimal strategies for training
  self-driving perception models in simulation,'' \emph{Advances in Neural
  Information Processing Systems}, vol.~34, pp. 1686--1699, 2021.

\bibitem{testing_add_14}
K.~Zhang, C.~Chang, W.~Zhong, S.~Li, Z.~Li, and L.~Li, ``A systematic solution
  of human driving behavior modeling and simulation for automated vehicle
  studies,'' \emph{IEEE Transactions on Intelligent Transportation Systems},
  2022.

\bibitem{testing_add_15}
M.~Johnson-Roberson, C.~Barto, R.~Mehta, S.~N. Sridhar, K.~Rosaen, and
  R.~Vasudevan, ``Driving in the matrix: Can virtual worlds replace
  human-generated annotations for real world tasks?'' \emph{arXiv preprint
  arXiv:1610.01983}, 2016.

\bibitem{testing_add_16}
S.~Kim, J.~Baek, J.~Park, G.~Kim, and S.~Kim, ``Instaformer: Instance-aware
  image-to-image translation with transformer,'' in \emph{Proceedings of the
  IEEE/CVF Conference on Computer Vision and Pattern Recognition}, 2022, pp.
  18\,321--18\,331.

\bibitem{human-machine_1}
M.~Hasenj{\"a}ger, M.~Heckmann, and H.~Wersing, ``A survey of personalization
  for advanced driver assistance systems,'' \emph{IEEE Transactions on
  Intelligent Vehicles}, vol.~5, no.~2, pp. 335--344, 2019.

\bibitem{human-machine_2}
Y.~Xing, C.~Lv, D.~Cao, and P.~Hang, ``Toward human-vehicle collaboration:
  Review and perspectives on human-centered collaborative automated driving,''
  \emph{Transportation Research Part C: Emerging Technologies}, vol. 128, p.
  103199, 2021.

\bibitem{human-machine_3}
W.~Wang and D.~Zhao, ``Evaluation of lane departure correction systems using a
  regenerative stochastic driver model,'' \emph{IEEE Transactions on
  Intelligent Vehicles}, vol.~2, no.~3, pp. 221--232, 2017.

\bibitem{human-machine_4}
W.~Chen, L.~Zhao, D.~Tan, Z.~Wei, K.~Xu, and Y.~Jiang, ``Human--machine shared
  control for lane departure assistance based on hybrid system theory,''
  \emph{Control Engineering Practice}, vol.~84, pp. 399--407, 2019.

\bibitem{human-machine_5}
M.~Althoff, S.~Maierhofer, and C.~Pek, ``Provably-correct and comfortable
  adaptive cruise control,'' \emph{IEEE Transactions on Intelligent Vehicles},
  vol.~6, no.~1, pp. 159--174, 2020.

\bibitem{human-machine_6}
R.~Utriainen, M.~P{\"o}ll{\"a}nen, and H.~Liimatainen, ``The safety potential
  of lane keeping assistance and possible actions to improve the potential,''
  \emph{IEEE Transactions on Intelligent Vehicles}, vol.~5, no.~4, pp.
  556--564, 2020.

\bibitem{human-machine_7}
S.~K. Lal and A.~Craig, ``A critical review of the psychophysiology of driver
  fatigue,'' \emph{Biological Psychology}, vol.~55, no.~3, pp. 173--194, 2001.

\bibitem{human-machine_8}
Y.~Xing, C.~Lv, H.~Wang, H.~Wang, Y.~Ai, D.~Cao, E.~Velenis, and F.-Y. Wang,
  ``Driver lane change intention inference for intelligent vehicles: framework,
  survey, and challenges,'' \emph{IEEE Transactions on Vehicular Technology},
  vol.~68, no.~5, pp. 4377--4390, 2019.

\bibitem{human-machine_9}
Z.~Hu, C.~Lv, P.~Hang, C.~Huang, and Y.~Xing, ``Data-driven estimation of
  driver attention using calibration-free eye gaze and scene features,''
  \emph{IEEE Transactions on Industrial Electronics}, vol.~69, no.~2, pp.
  1800--1808, 2021.

\bibitem{human-machine_10}
G.~Marquart, C.~Cabrall, and J.~de~Winter, ``Review of eye-related measures of
  drivers’ mental workload,'' \emph{Procedia Manufacturing}, vol.~3, pp.
  2854--2861, 2015.

\bibitem{human-machine_11}
S.~Zepf, J.~Hernandez, A.~Schmitt, W.~Minker, and R.~W. Picard, ``Driver
  emotion recognition for intelligent vehicles: A survey,'' \emph{ACM Computing
  Surveys (CSUR)}, vol.~53, no.~3, pp. 1--30, 2020.

\bibitem{human-machine_12}
A.~Kashevnik, R.~Shchedrin, C.~Kaiser, and A.~Stocker, ``Driver distraction
  detection methods: A literature review and framework,'' \emph{IEEE Access},
  vol.~9, pp. 60\,063--60\,076, 2021.

\bibitem{human-machine_16}
N.~Du, F.~Zhou, E.~M. Pulver, D.~M. Tilbury, L.~P. Robert, A.~K. Pradhan, and
  X.~J. Yang, ``Predicting driver takeover performance in conditionally
  automated driving,'' \emph{Accident Analysis \& Prevention}, vol. 148, p.
  105748, 2020.

\bibitem{human-machine_17}
B.~W. Weaver and P.~R. DeLucia, ``A systematic review and meta-analysis of
  takeover performance during conditionally automated driving,'' \emph{Human
  Factors}, p. 0018720820976476, 2020.

\bibitem{human-machine_23}
P.~Jing, G.~Xu, Y.~Chen, Y.~Shi, and F.~Zhan, ``The determinants behind the
  acceptance of autonomous vehicles: A systematic review,''
  \emph{Sustainability}, vol.~12, no.~5, p. 1719, 2020.

\bibitem{human-machine_24}
C.~Olaverri-Monreal, ``Promoting trust in self-driving vehicles,'' \emph{Nature
  Electronics}, vol.~3, no.~6, pp. 292--294, 2020.

\bibitem{human-machine_25}
J.~K. Choi and Y.~G. Ji, ``Investigating the importance of trust on adopting an
  autonomous vehicle,'' \emph{International Journal of Human-Computer
  Interaction}, vol.~31, no.~10, pp. 692--702, 2015.

\bibitem{human-machine_26}
M.~Dikmen and C.~Burns, ``Trust in autonomous vehicles: The case of tesla
  autopilot and summon,'' in \emph{2017 IEEE International Conference on
  Systems, Man, and Cybernetics (SMC)}.\hskip 1em plus 0.5em minus 0.4em\relax
  IEEE, 2017, pp. 1093--1098.

\bibitem{human-machine_27}
J.~Ayoub, F.~Zhou, S.~Bao, and X.~J. Yang, ``From manual driving to automated
  driving: A review of 10 years of autoui,'' in \emph{Proceedings of the 11th
  International Conference on Automotive User Interfaces and Interactive
  Vehicular Applications}, 2019, pp. 70--90.

\bibitem{human-machine_28}
W.~Liu, Q.~Li, Z.~Wang, W.~Wang, C.~Zeng, and B.~Cheng, ``A literature review
  on additional semantic information conveyed from driving automation systems
  to drivers through advanced in-vehicle {HMI} just before, during, and right
  after takeover request,'' \emph{International Journal of Human--Computer
  Interaction}, pp. 1--21, 2022.

\bibitem{human-machine_29}
I.~Politis, P.~Langdon, M.~Bradley, L.~Skrypchuk, A.~Mouzakitis, P.~J.
  Clarkson, and N.~A. Stanton, ``Designing autonomy in cars: A survey and two
  focus groups on driving habits of an inclusive user group, and group
  attitudes towards autonomous cars,'' in \emph{Designing Interaction and
  Interfaces for Automated Vehicles}.\hskip 1em plus 0.5em minus 0.4em\relax
  CRC Press, 2021, pp. 41--54.

\bibitem{human-machine_30}
O.~Benderius, C.~Berger, and V.~M. Lundgren, ``The best rated human--machine
  interface design for autonomous vehicles in the 2016 grand cooperative
  driving challenge,'' \emph{IEEE Transactions on Intelligent Transportation
  Systems}, vol.~19, no.~4, pp. 1302--1307, 2017.

\bibitem{human-machine_31}
K.~Bengler, M.~Rettenmaier, N.~Fritz, and A.~Feierle, ``From {HMI} to {HMI}s:
  Towards an {HMI} framework for automated driving,'' \emph{Information},
  vol.~11, no.~2, p.~61, 2020.

\bibitem{human-machine_32}
F.~Wulf, M.~Rimini-D{\"o}ring, M.~Arnon, and F.~Gauterin, ``Recommendations
  supporting situation awareness in partially automated driver assistance
  systems,'' \emph{IEEE Transactions on Intelligent Transportation Systems},
  vol.~16, no.~4, pp. 2290--2296, 2014.

\bibitem{human-machine_33}
N.~T. Richardson, C.~Lehmer, M.~Lienkamp, and B.~Michel, ``Conceptual design
  and evaluation of a human machine interface for highly automated truck
  driving,'' in \emph{2018 IEEE Intelligent Vehicles Symposium (IV)}.\hskip 1em
  plus 0.5em minus 0.4em\relax IEEE, 2018, pp. 2072--2077.

\bibitem{IC_1}
E.~Ohn-Bar and M.~M. Trivedi, ``Looking at humans in the age of self-driving
  and highly automated vehicles,'' \emph{IEEE Transactions on Intelligent
  Vehicles}, vol.~1, no.~1, pp. 90--104, 2016.

\bibitem{survey_3}
R.~Hussain and S.~Zeadally, ``Autonomous cars: Research results, issues, and
  future challenges,'' \emph{IEEE Communications Surveys \& Tutorials},
  vol.~21, no.~2, pp. 1275--1313, 2018.

\bibitem{IC_4}
M.~N. Ahangar, Q.~Z. Ahmed, F.~A. Khan, and M.~Hafeez, ``A survey of autonomous
  vehicles: Enabling communication technologies and challenges,''
  \emph{Sensors}, vol.~21, no.~3, p. 706, 2021.

\bibitem{IC_5}
J.~Yang, S.~Xing, Y.~Chen, R.~Qiu, C.~Hua, and D.~Dong, ``A comprehensive
  evaluation model for the intelligent automobile cockpit comfort,''
  \emph{Scientific Reports}, vol.~12, no.~1, pp. 1--9, 2022.

\bibitem{IC_6}
R.~Tan, W.~Li, F.~Hu, X.~Xiao, S.~Li, Y.~Xing, H.~Wang, and D.~Cao, ``Motion
  sickness detection for intelligent vehicles: A wearable-device-based
  approach,'' in \emph{2022 IEEE 25th International Conference on Intelligent
  Transportation Systems (ITSC)}.\hskip 1em plus 0.5em minus 0.4em\relax IEEE,
  2022, pp. 4355--4362.

\bibitem{IC_7}
O.~Carsten and M.~H. Martens, ``How can humans understand their automated cars?
  hmi principles, problems and solutions,'' \emph{Cognition, Technology \&
  Work}, vol.~21, no.~1, pp. 3--20, 2019.

\end{thebibliography}
